\documentclass[letterpaper,times]{IONconf}

\usepackage{titling}
\usepackage{hyperref} 
\usepackage{url} 
\usepackage{fancyhdr}
\usepackage{courier} 
\usepackage{amsmath}
\usepackage{amsfonts}
\usepackage[euler]{textgreek}
\usepackage{bbm} 
\usepackage[]{graphicx}
\usepackage{float}
\usepackage{lipsum}
\usepackage{subcaption}
\usepackage{dsfont}
\usepackage{mathtools}
\usepackage{amssymb}
\usepackage{siunitx} 
\usepackage{xcolor}
\usepackage[ruled,vlined,linesnumbered]{algorithm2e}
\IncMargin{0.5cm} 

\usepackage{outlines}
\usepackage{enumitem}
\setenumerate[2]{label=\alph*.}
\setenumerate[3]{label=\roman*.}

\usepackage{nopageno}
\usepackage[backend=biber,style=apa, natbib=true]{biblatex}
\newcommand{\new}[1]{{\color{black}{#1}}}

\newcommand{\zeros}{\mathbf{0}}

\newcommand{\eye}{{I}}
\newcommand{\zeromatrix}{O}

\newcommand{\state}{\mathbf{x}}
\newcommand{\control}{\mathbf{u}}
\newcommand{\controlmatrix}{B}

\newcommand{\measurement}{\mathbf{z}}
\newcommand{\motionerr}{\mathbf{w}}
\newcommand{\motioncov}{Q}
\newcommand{\measurementerr}{\mathbf{v}}
\newcommand{\measurementcov}{R}
\newcommand{\statecov}{P}
\newcommand{\stateestcov}{\Lambda}
\newcommand{\totalstatecov}{\Sigma}
\newcommand{\kalmangain}{G}
\newcommand{\feedbackgain}{K}
\newcommand{\valuefunc}{V}
\newcommand{\adjacency}{A}
\newcommand{\laplacian}{L}
\newcommand{\degreematrix}{D}
\newcommand{\degreevector}{\mathbf{d}}

\newcommand{\AC}{\lambda_2}
\newcommand{\AClimit}{\epsilon}
\newcommand{\Dt}{\Delta t}
\newcommand{\commrange}{\rho}
\newcommand{\losrange}{d^\beta}
\newcommand{\collrange}{d^\gamma}
\newcommand{\edgeweight}{a}
\newcommand{\rangefactor}{\alpha}
\newcommand{\LOSfactor}{\beta}
\newcommand{\collisionfactor}{\gamma}
\newcommand{\distmeasure}{\bar{l}}

\newcommand{\eigenvalue}{\lambda}
\newcommand{\eigenvector}{\mathbf{e}}
\newcommand{\gradient}[2]{ \frac{\partial #1}{\partial #2} }
\newcommand{\degree}{d}
\newcommand{\diag}{\texttt{diag}}
\newcommand{\N}{n}

\title{Decentralized Connectivity Maintenance for Multi-robot Systems Under Motion and Sensing Uncertainties}

\author{
    Akshay Shetty, Timmy Hussain and Grace Gao \\ \textit{Stanford~University}
}

\begin{document}
\date{}
\maketitle

\section*{biography}



\biography{Akshay~Shetty}{is a postdoctoral scholar in the Department of Aeronautics and Astronautics at Stanford University. He obtained his Ph.D. in Aerospace Engineering from University of Illinois at Urbana-Champaign, and his B.Tech. in Aerospace Engineering from Indian Institute of Technology Bombay. His research is in the field of safe navigation of ground and aerial autonomous vehicles, focusing on trajectory planning, sensor fusion and learning-based algorithms.}

\biography{Timmy Hussain}{is an M.S Candidate in the Department of Aeronautics and Astronautics at Stanford University. He received his B.S in Aerospace Engineering from Massachusetts Institute of Technology. His research interests include the control and navigation of autonomous systems.}

\biography{Grace~Gao}{is an assistant professor in the Department of Aeronautics and Astronautics at Stanford University. Before joining Stanford University, she was an assistant professor at University of Illinois at Urbana-Champaign. She obtained her Ph.D. degree at Stanford University. Her research is on robust and secure positioning, navigation and timing with applications to manned and unmanned aerial vehicles, robotics, and power systems.}
\section*{Abstract}

Communication connectivity is desirable for safe and efficient operation of multi-robot systems. While decentralized algorithms for connectivity maintenance have been explored in recent literature, the majority of these works do not account for robot motion and sensing uncertainties. These uncertainties are inherent in practical robots and result in robots deviating from their desired positions which could potentially result in a loss of connectivity. In this paper we present a Decentralized Connectivity Maintenance algorithm accounting for robot motion and sensing Uncertainties (DCMU). We first propose a novel weighted graph definition for the multi-robot system that accounts for the aforementioned uncertainties along with realistic connectivity constraints such as line-of-sight connectivity and collision avoidance. Next we design a decentralized gradient-based controller for connectivity maintenance where we derive the gradients of our weighted graph edge weights required for computing the control. Finally, we perform multiple simulations to validate the connectivity maintenance performance of our DCMU algorithm under robot motion and sensing uncertainties and show an improvement compared to previous work.


\section{Introduction}\label{sec:intro}

Multi-robot systems are increasingly being used for various tasks such as exploration, target tracking, and search and rescue~\citep{rizk2019cooperative,cortes2017coordinated}. One of the primary advantages of multi-robot systems is their ability to coordinate using inter-robot communication, which allows them to execute complex tasks in a safe and efficient manner~\citep{alanwar2019distributed,bhamidipati2019locating,park2018robust}. Thus, it is highly desirable to maintain communication connectivity within the multi-robot system.

Existing literature has explored \textit{decentralized} algorithms for connectivity maintenance due to their communication efficiency and scalability properties~\citep{khateri2019comparison}. These works typically represent the system as a weighted graph, where each node represents a robot and each edge represents the communication connection between two robots~\citep{yang2010decentralized,Sabattini2013,robuffo2013passivity}. The edge weights between robots are formulated as a function of the robot positions subject to constraints such as maximum communication range, line-of-sight communication and collision avoidance (both inter-robot and obstacles). One commonly used metric for the connectivity of the system is algebraic connectivity, which depends on the graph edge weights as explained later in Section~\ref{sec:preliminaries}. A value greater than zero indicates that the system is connected. Thus, in order to maintain connectivity within the system, previous works~\citep{yang2010decentralized,Sabattini2013,robuffo2013passivity} derive a decentralized gradient-based controller to maintain the algebraic connectivity above a specified lower limit.

While previous works present efficient decentralized algorithms for connectivity maintenance, they assume that the robot positions are deterministic and do not explicitly account for robot motion and sensing uncertainties. Here, motion uncertainties refer to the errors between the actual robot motion and a mathematical motion model, whereas sensing uncertainties refer to errors in measurements such as errors in localization measurements. These motion and sensing uncertainties, which are inherent in practical robots, result in robots deviating from their desired nominal positions. These deviations, if not accounted for, can potentially result in a loss of connectivity in the multi-robot system. Thus, it is important to account for robot motion and sensing uncertainties while designing connectivity maintenance algorithms.

In our prior work~\citep{shetty2020connectivity} we define a weighted graph to represent the connectivity of the multi-robot system while accounting for deviations due to motion and sensing uncertainties. We then use a distributed trajectory planner to maintain connectivity within the system. However, our planner in~\citep{shetty2020connectivity} requires a centralized communication setup, i.e., each robot communicates with all other robots in the system in a single planning iteration. While such communication is possible (perhaps via multi-hop connections), in practice it introduces additional communication delays and does not scale favorably to large systems. Thus, decentralized algorithms, as proposed in~\citep{yang2010decentralized,Sabattini2013,robuffo2013passivity}, are desirable since they require a robot to communicate only with its immediate neighboring robots in a single planning iteration. Additionally, the weighted graph in~\citep{shetty2020connectivity} assumes a simplistic connectivity model subject to only a maximum communication range constraint, as opposed to the more realistic line-of-sight communication and collision avoidance constraints in~\citep{robuffo2013passivity}.

\begin{table}[t]
\centering
\begin{tabular}{ |>{\centering\arraybackslash}m{7cm}|>{\centering\arraybackslash}m{2cm}|>{\centering\arraybackslash}m{3cm}|>{\centering\arraybackslash}m{3cm}| } 
 \hline
   & Decentralized & Motion and sensing uncertainties & Line-of-sight and collision avoidance \\ 
 \hline
 \citep{yang2010decentralized}\citep{Sabattini2013} \ \ \ \ \ \ \ \ \ \ \ \ \  \citep{gasparri2017bounded}\citep{siligardi2019robust} & \checkmark &  &  \\ 
 \hline
 \citep{robuffo2013passivity} & \checkmark & & \checkmark\\ 
 \hline
 \citep{shetty2020connectivity} & & \checkmark & \\
 \hline
 \textbf{This work} & \checkmark & \checkmark & \checkmark \\
 \hline
\end{tabular}
\vspace{0.1cm}
\caption{Comparison of connectivity maintenance works in terms of their decentralized nature, explicit accounting of motion and sensing uncertainties, and considering realistic constraints such as line-of-sight communication and collision avoidance.}
\label{table:comparison}
\end{table}

In this work we present a decentralized algorithm (referred to as DCMU) for connectivity maintenance of multi-robot systems while accounting for motion and sensing uncertainties. A decentralized gradient-based controller is implemented in order to maintain system connectivity subject to constraints of maximum communication range, line-of-sight communication and collision avoidance. Table~\ref{table:comparison} outlines the contributions of this work in comparison to previous connectivity maintenance works. This paper is based on our work in~\citep{shetty2021decentralized}. The main contributions of this work are listed as follows:
\begin{enumerate}

    \item We propose a weighted graph that accounts for deviations in robot positions arising due to \textbf{motion and sensing uncertainties}. Compared to our prior work~\citep{shetty2020connectivity}, we include more realistic constraints of \textbf{line-of-sight} communication and \textbf{collision avoidance} (both inter-robot and obstacles) in addition to a maximum communication range.
    
    \item Next, we derive a \textbf{decentralized gradient-based controller} in order to maintain the algebraic connectivity of the proposed weighted graph above a specified lower limit. Here we derive the gradients of the edge weights of our weighted graph that are required for computing the control. These edge weights account for deviations in robot positions instead of assuming deterministic robot positions as in~\citep{yang2010decentralized,Sabattini2013}.
    
    \item Finally, we present extensive simulation results to evaluate our DCMU algorithm under motion and sensing uncertainties. We compare our connectivity maintenance performance with~\citep{Sabattini2013} and validate our algorithm on a high-fidelity simulator~\citep{shah2018airsim}.

\end{enumerate}

The remainder of the paper is organized as follows. We first introduce relevant background from graph theory in Section~\ref{sec:preliminaries}. Next, in Section~\ref{sec:problem_formulation} we describe the models used for the robot motion and sensing uncertainties, and formulate the connectivity maintenance problem. Section~\ref{sec:dcm_algorithm} provides details of our DCMU algorithm including the weighted graph definition and the gradient-based controller. Finally, in Section~\ref{sec:results} we present simulation results validating the connectivity maintenance performance of our algorithm.
\section{Preliminaries: Graph Theory}\label{sec:preliminaries}

A multi-robot system can be represented as an undirected graph, where each node represents a robot and each edge represents the communication connection between two robots. Let $\N$ be the number of nodes in the graph. The adjacency matrix $\adjacency$ of the graph is defined as a $\N \times \N$ matrix where $\edgeweight_{ij} \in [0,1]$ represents the edge weight between two nodes $i$ and $j$, with $\edgeweight_{ii} = 0$~\citep{grone1990laplacian}. The degree of a node is defined as $\degree_i = \sum_{j=1}^{\N} \edgeweight_{ij}$. The vector of node degrees $\degreevector = [\degree_1, \hdots, \degree_n]$ is then used to define the degree matrix $\degreematrix$ of the graph as $\degreematrix = \diag(\degreevector)$. Given matrices $\adjacency$ and $\degreematrix$ the Laplacian matrix $\laplacian$ of the graph is defined as $\laplacian = \degreematrix - \adjacency$~\citep{grone1990laplacian}.

\begin{figure}[t]
  \centering
  \subfloat[]{\includegraphics[width=0.3\linewidth]{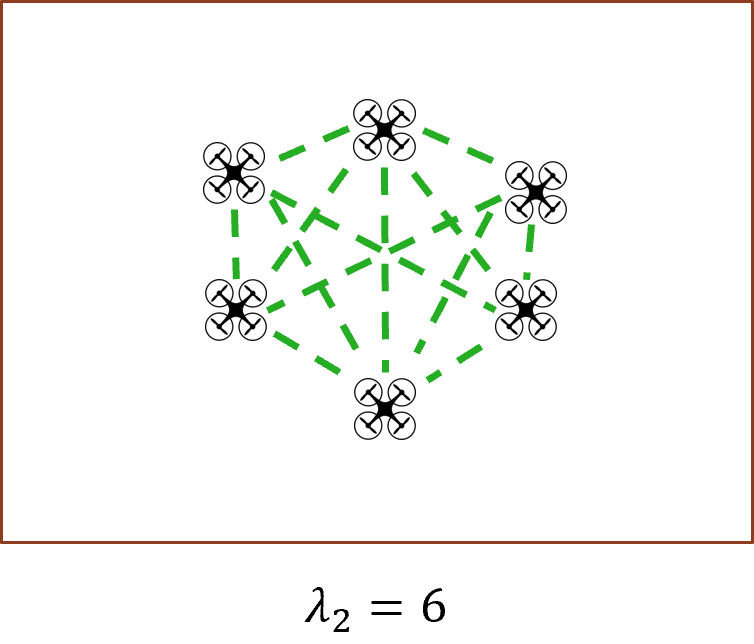}}\hfill
  \subfloat[]{\includegraphics[width=0.3\linewidth]{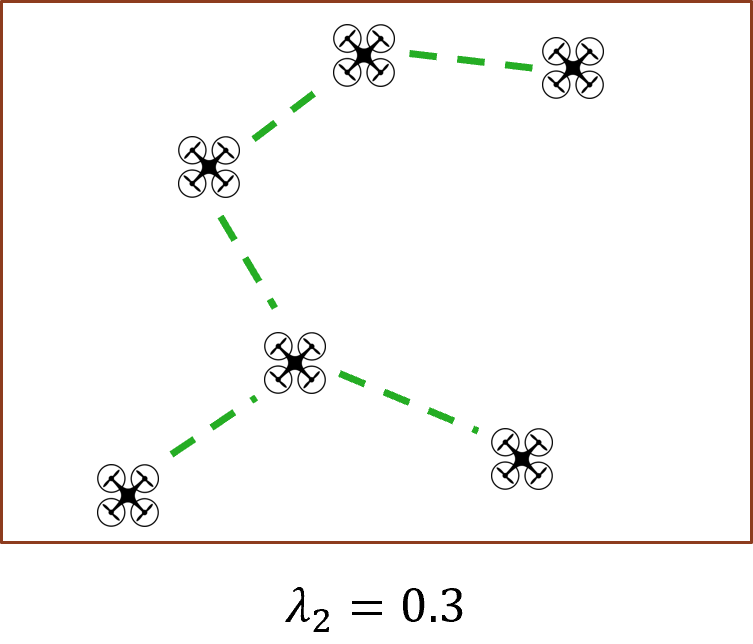}}\hfill
  \subfloat[]{\includegraphics[width=0.3\linewidth]{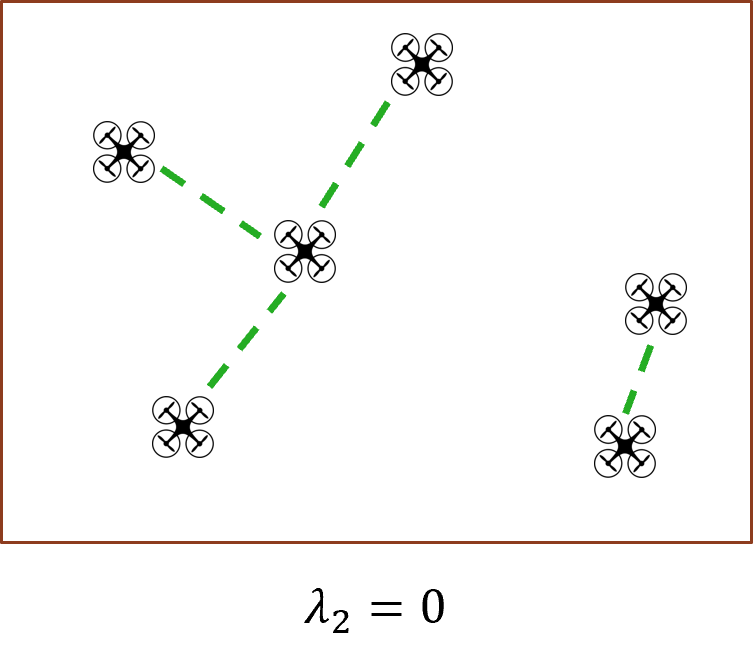}}\hfill
  \caption{The algebraic connectivity $\AC$ of a graph varies from the number of nodes (fully-connected graph) to zero (disconnected graph). $\AC > 0$ implies that the graph is connected.}
  \label{fig:bacground-algebraic-connectivity}
\end{figure}

Let $\eigenvalue_1~\leq~ \AC \leq \dots \leq \eigenvalue_\N$ be the eigenvalues of $\laplacian$. The algebraic connectivity of the graph, also known as the Fiedler value, is defined as the second-smallest eigenvalue of $\laplacian$, i.e., $\AC$. As mentioned in Section~\ref{sec:intro}, algebraic connectivity is a commonly used metric to represent the connectivity of multi-robot systems. It varies from zero (if the graph is disconnected) to the number nodes in the graph (if the graph is fully connected), i.e. $0 \leq \AC \leq \N$. Fig.~\ref{fig:bacground-algebraic-connectivity} illustrates the algebraic connectivity for different configurations of a graph. Thus, $\AC$ remains greater than $0$ as long as there exists a (potentially multi-hop) communication path between any two robots in the system.

\section{Problem Formulation}\label{sec:problem_formulation}

\subsection{Robot description}
\label{subsec:robot-description}
For each robot $i$ in the multi-robot system, we consider a discrete-time single-integrator motion model:
\begin{equation}
   \label{eqn:motion-model}
    \state_{i,t} = \state_{i,t-1} + \controlmatrix \control_{i,t-1} + \motionerr_{i,t},
\end{equation}
where $t$ represents the time instant, $\state_{i,t}$ is the state vector representing the robot position, $\control_{i,t}$ is the velocity control input, $\controlmatrix$ is the control input matrix, and $\motionerr_{i,t}$ is a zero-mean Gaussian-distributed error vector with covariance matrix $\motioncov_{i,t}$, i.e., $\motionerr_{i,t} \sim \mathcal{N}(\zeros, \motioncov_{i,t})$. The control input matrix in Equation~(\ref{eqn:motion-model}) is set as $\controlmatrix = (\Dt) \eye$, where $\Dt$ is the discrete time-step. Note that while this is a simplified motion model, it can still be used to control real robots as demonstrated in previous works on aerial robots~\citep{soukieh2009obstacle,lee2013semiautonomous}. For the sensing model, we assume each robot $i$ obtains position measurements:
\begin{equation}
    \label{eqn:sensing-model}
    \measurement_{i,t} = \state_{i,t} + \measurementerr_{i,t},
\end{equation}
where $\measurement_{i,t}$ is the measurement vector representing position measurements and $\measurementerr_{i,t}$ is a zero-mean Gaussian-distributed error vector with covariance matrix $\measurementcov_{i,t}$, i.e., $\measurementerr_{i,t} \sim \mathcal{N}(\zeros, \measurementcov_{i,t})$. Given the linear motion and sensing models with Gaussian uncertainties, we set each robot to use a Kalman Filter (KF) for state estimation. The prediction step of the KF is performed as:
\begin{align}
    \label{eqn:ekf-pred-1}
    \bar{\state}_{i,t} &= \hat{\state}_{i,t-1} + \controlmatrix \control_{i,t-1}, \\
    \label{eqn:ekf-pred-2}
    \bar{\statecov}_{i,t} &= \statecov_{i,t-1} + \motioncov_{i,t},
\end{align}
where $\statecov_{i,t}$ is the state estimation covariance matrix such that $\state_{i,t} \sim \mathcal{N}(\hat{\state}_{i,t}, \statecov_{i,t})$. The correction step of the KF is performed as:
\begin{align}
    \label{eqn:ekf-update-1}
    \kalmangain_{i,t} &= \bar{\statecov}_{i,t} (\bar{\statecov}_{i,t} + \measurementcov_{i,t})^{-1}, \\
    \label{eqn:ekf-update-2}
    \hat{\state}_{i,t} &= \bar{\state}_{i,t} + \kalmangain_{i,t}(\measurement_{i,t} - \bar{\state}_{i,t} ), \\
    \label{eqn:ekf-update-3}
    \statecov_{i,t} &= \bar{\statecov}_{i,t} - \kalmangain_{i,t} \bar{\statecov}_{i,t},
\end{align}
where $\kalmangain_{i,t}$ is the Kalman gain matrix. In order to track a desired nominal state $\check{\state}_{i,t}$, we assume that each robot uses linear feedback control, where the total control input $\control_{i,t}$ is of the form:
\begin{equation}
    \label{eqn:total-control}
    \control_{i,t} = \check{\control}_{i,t} - \check{\feedbackgain}_i( \hat{\state}_{i,t} - \check{\state}_{i,t} ),
\end{equation}
where $\check{\feedbackgain}_i$ is the feedback control gain which can be designed using methods such as classical control theory or LQR design~\citep{doyle2013feedback}, and $\check{\control}_{i,t}$ is the nominal control input which relates to the nominal states as:
\begin{equation}
    \label{eqn:nominal-state}
    \check{\state}_{i,t} = \check{\state}_{i,t-1} + \controlmatrix \check{\control}_{i,t-1}.
\end{equation}
Later in Section~\ref{subsec:connectivity-maintenance} we discuss how the nominal control input is obtained for different robots in the multi-robot system.

Given the above setup for each robot $i$, the distribution of the robot's true state $\state_{i,t}$ about its nominal state $\check{\state}_{i,t}$ can be obtained using the method described in~\citep{bry2011rapidly}. The authors derive the following Gaussian distribution for the robot's true state:
\begin{equation}
\label{eqn:total-state-cov-dist}
\state_{i,t} \sim \mathcal{N}( \check{\state}_{i,t}, \totalstatecov_{i,t} )
\end{equation}
where $\totalstatecov_{i,t} = \statecov_{i,t} + \stateestcov_{i,t}$. Here $\statecov_{i,t}$ is the state estimation covariance matrix from Equation~(\ref{eqn:ekf-update-3}), and $\stateestcov_{i,t}$ can be obtained iteratively as follows~\citep{bry2011rapidly}:
\begin{equation}
    \label{eqn:state-est-cov}
    \stateestcov_{i,t} = (\eye - \controlmatrix \check{\feedbackgain}_i) \stateestcov_{i,t-1} (\eye - \controlmatrix \check{\feedbackgain}_i)^\top + \kalmangain_{i,t} \bar{\statecov}_{i,t},
\end{equation}
where $\eye$ is an identity matrix and $\stateestcov_{i,0} = \zeromatrix$. In essence, the distribution $\state_{i,t} \sim \mathcal{N}( \check{\state}_{i,t}, \totalstatecov_{i,t} )$ in Equation~(\ref{eqn:total-state-cov-dist}) captures the deviations of the robot's position from its desired nominal position that arise due to the motion and sensing uncertainties.

\subsection{Multi-robot system: Connectivity maintenance}
\label{subsec:connectivity-maintenance}

We consider the multi-robot system to be comprised of two types of robots: leader robots and follower robots. For the leader robots we assume that the nominal control inputs in Equation~(\ref{eqn:nominal-state}) are available from a high-level planner such as from an exploration or a search and rescue strategy~\citep{burgard2005coordinated,baxter2007multi}. On the other hand, the objective of follower robots is to maintain connectivity within the multi-robot system. Our DCMU algorithm presented in Section~\ref{sec:dcm_algorithm} focuses on deriving the nominal control inputs for these follower robots.

In practice the communication connectivity between two robots can be considered to be binary, i.e., the robots are either connected if certain constraints are satisfied (such as minimum signal strength, bandwidth, etc.) or they are disconnected. We consider any two robots $i$ and $j$ in the multi-robot system to be connected if the following constraints are satisfied:
\begin{enumerate}
    \item Communication range constraint: The distance between the robots is within a maximum communication range $\commrange$, i.e., $\left \| \state_{i,t} - \state_{j,t} \right \|_2 \leq \commrange$.
    \item Line-of-sight constraint: The robots maintain a direct line-of-sight, i.e., the line segment connecting $\state_{i,t}$ and $\state_{j,t}$ is not obstructed by obstacles in the environment.
    \item Collision avoidance constraint: The robots are not colliding with any obstacles or other robots in the system. This constraint helps ensure that the communication equipment is not possibly damaged due a collision.
\end{enumerate}
Let $\bar{\edgeweight}_{ij} = \{ 0,1 \}$ represent the binary connectivity between robots $i$ and $j$, i.e., $\bar{\edgeweight}_{ij} = 1$ if the robots are connected (the above constraints are satisfied) and $\bar{\edgeweight}_{ij} = 0$ otherwise (any of the above constraints is not satisfied). Note that by definition $\bar{\edgeweight}_{ii} = 0$. These $\bar{\edgeweight}$ can then be used to obtain the algebraic connectivity $\bar{\eigenvalue}_2$ (as explained in Section~\ref{sec:preliminaries}) of the graph which represents the connectivity of the multi-robot system based on the actual robot positions $\state_{i,t}$. Thus, $\bar{\eigenvalue}_2 > 0$ implies that the multi-robot system is connected.

\begin{figure}[t]
  \centering
  \subfloat{\includegraphics[width=0.8\linewidth]{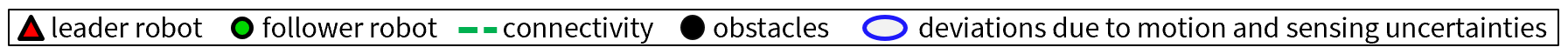} \vspace{0.2cm}}\addtocounter{subfigure}{-1}\hfill \\
  \subfloat{\includegraphics[width=0.8\linewidth]{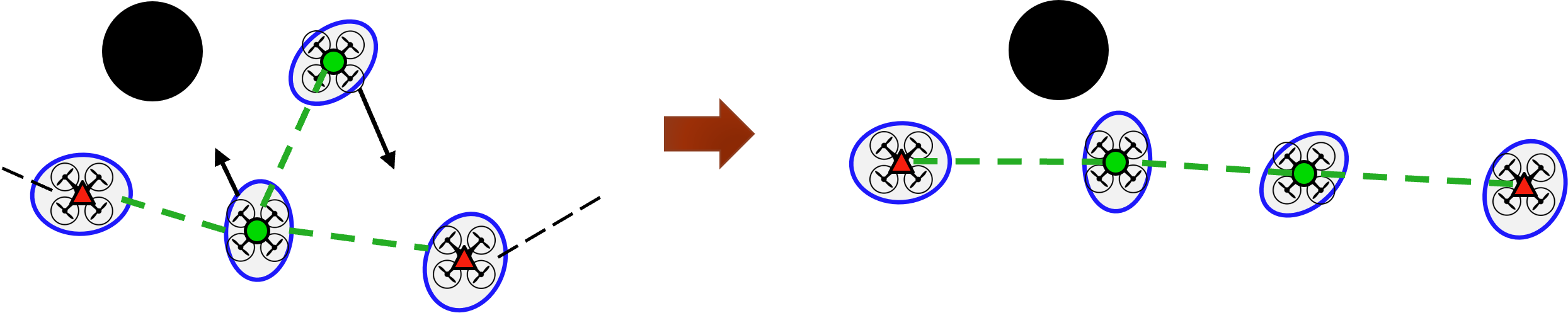}}\hfill
  \caption{Illustration of the problem definition for our DCMU algorithm. Given a multi-robot system with leader robots and follower robots where the leader robots are controlled by some high-level planner (black-dashed lines), the objective of our DCMU algorithm is to derive nominal control inputs (black arrows) for the follower robots such that communication connectivity is maintained within the multi-robot system. Note that our algorithm must account for deviations in robot positions due to motion and sensing uncertainties.}
  \label{fig:problem-formulation}
\end{figure}

We define the problem for our DCMU algorithm as: given a multi-robot system as described in this section, derive nominal control inputs ($\check{\control}_{i,t}$ in Equation~(\ref{eqn:nominal-state})) for the follower robots such that $\bar{\eigenvalue}_2$ is always above a specified lower limit $\AClimit$, i.e., $\bar{\eigenvalue}_2 > \AClimit$. Fig.~\ref{fig:problem-formulation} illustrates the defined problem.

\section{Proposed Connectivity Maintenance Algorithm: DCMU}\label{sec:dcm_algorithm}

During execution, each robot $i$ in the multi-robot system deviates from its desired nominal position $\check{\state}_{i,t}$ due to motion and sensing uncertainties. Previous connectivity maintenance works~\citep{yang2010decentralized,Sabattini2013,robuffo2013passivity,gasparri2017bounded,siligardi2019robust} that do not account for these uncertainties essentially assume that the robots are at their nominal positions, i.e., $\state_{i,t} = \check{\state}_{i,t}$. In contrast, our DCMU algorithm accounts for the deviations of $\state_{i,t}$ from $\check{\state}_{i,t}$ as modeled by Equation~(\ref{eqn:total-state-cov-dist}).

In this section we provide the details of our DCMU algorithm. We first define a weighted graph to represent the connectivity of the multi-robot system where the edge weights account for deviations due to motion and sensing uncertainties. Next, we implement a decentralized power iteration method for each robot to estimate connectivity information of the weighted graph. Finally, each follower robot uses a gradient-based controller to obtain the nominal control input $\check{\control}_{i,t}$ for maintaining connectivity of the weighted graph. Here we derive the expressions for the gradients of our weighted graph edge weights, which are required to compute the nominal control input $\check{\control}_{i,t}$. While for simplicity we drop the time notation in the remainder of this section, the nominal control input computed by our DCMU algorithm is applicable for all time instants $t$.

\begin{figure}[t]
  \centering
  \subfloat[]{\includegraphics[width=0.4\linewidth]{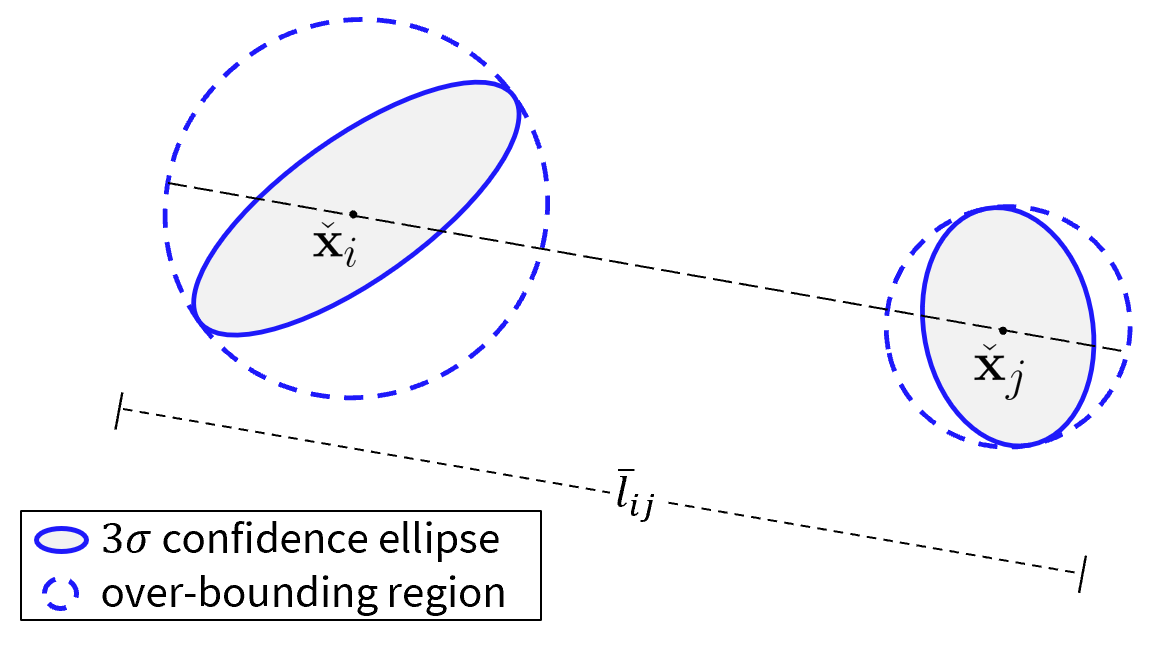}}\hfill
  \subfloat[]{\includegraphics[width=0.4\linewidth]{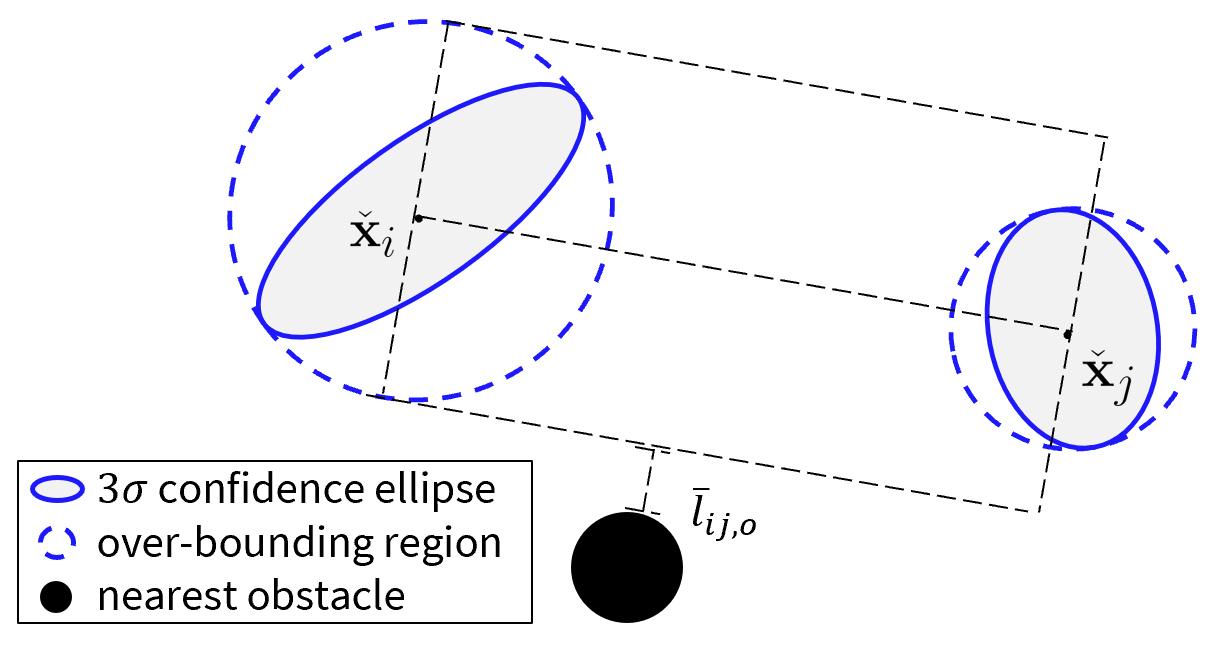}}\hfill \\
  \subfloat[]{\includegraphics[width=0.4\linewidth]{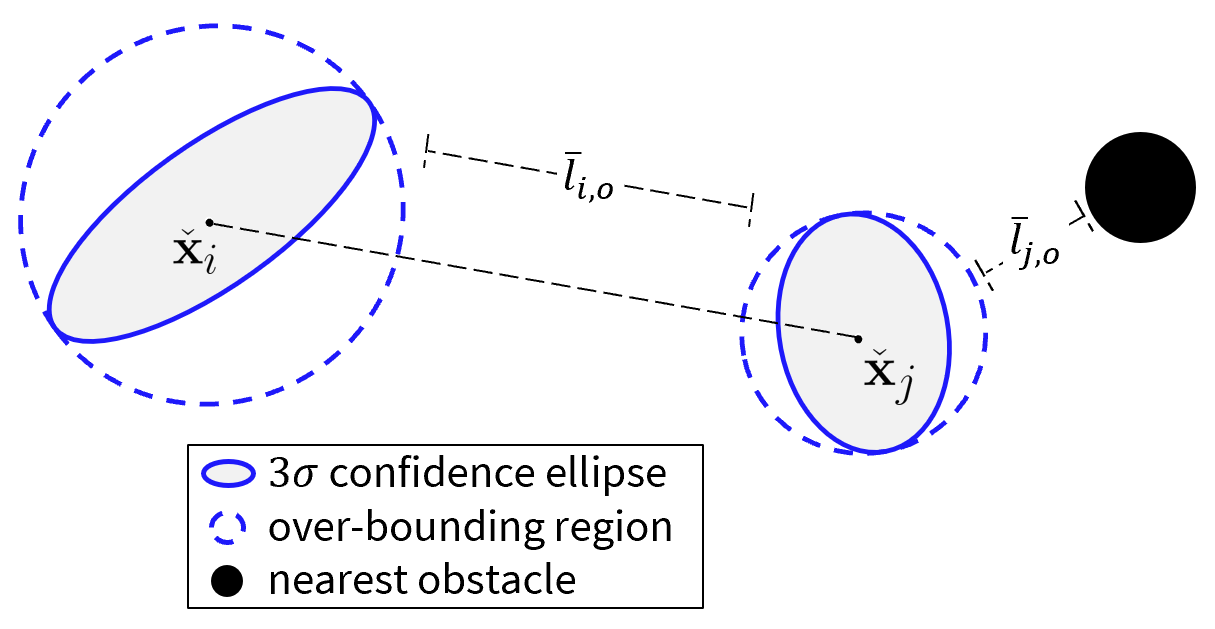}}\hfill
  \caption{Conservative measures used to define our weighted graph in Section~\ref{subsec:edge-weights} that accounts for robot deviations due to motion and sensing uncertainties around their nominal positions $\check{\state}_i$ and $\check{\state}_j$. (a) $\distmeasure_{ij}$: A conservative measure of the distance between two robots used for the communication range factor in Section~\ref{subsec:edge-weights}.1. (b) $\distmeasure_{ij,o}$: A conservative measure of distance between the line-of-sight and the closest obstacle used for the line-of-sight factor in Section~\ref{subsec:edge-weights}.2. (c) $\distmeasure_{i,o},\distmeasure_{j,o}$: Conservative measures of the closest collision points for two robots. In this illustration the closest collision point for robot $i$ is robot $j$, whereas for robot $j$ it is the obstacle.}
  \label{fig:weighted-graph-lenghts}
\end{figure}
\begin{figure}[t!]
  \centering
  \hspace{1.5cm}\subfloat[]{\includegraphics[width=0.25\linewidth]{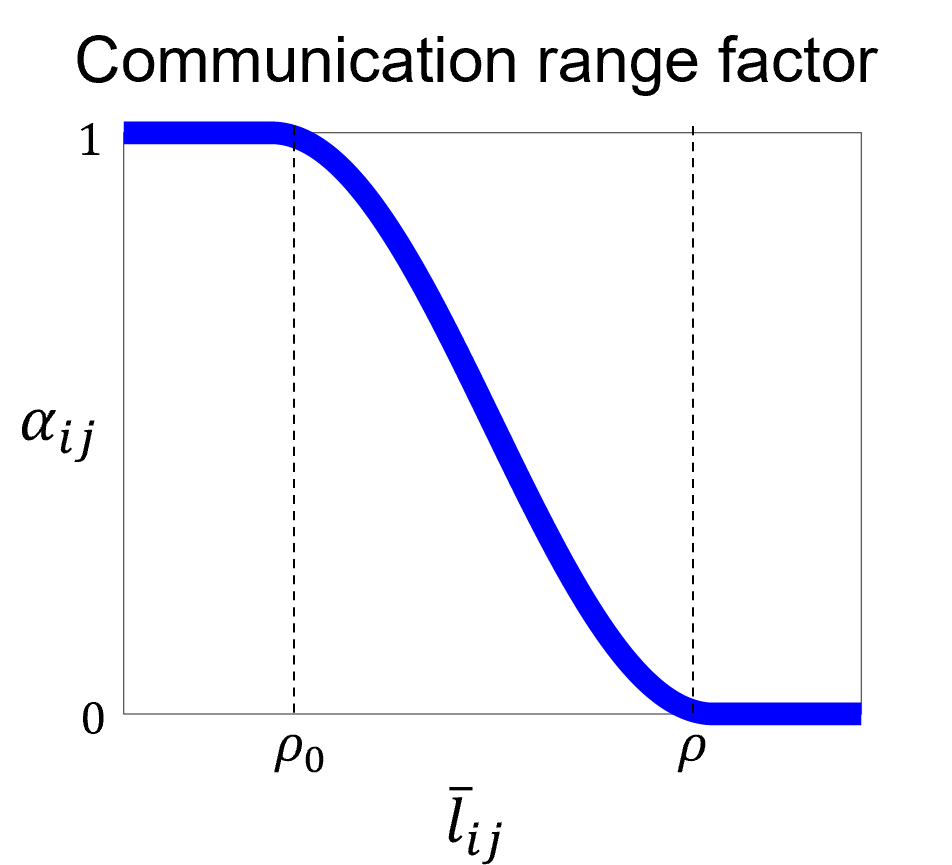}}\hfill
  \subfloat[]{\includegraphics[width=0.25\linewidth]{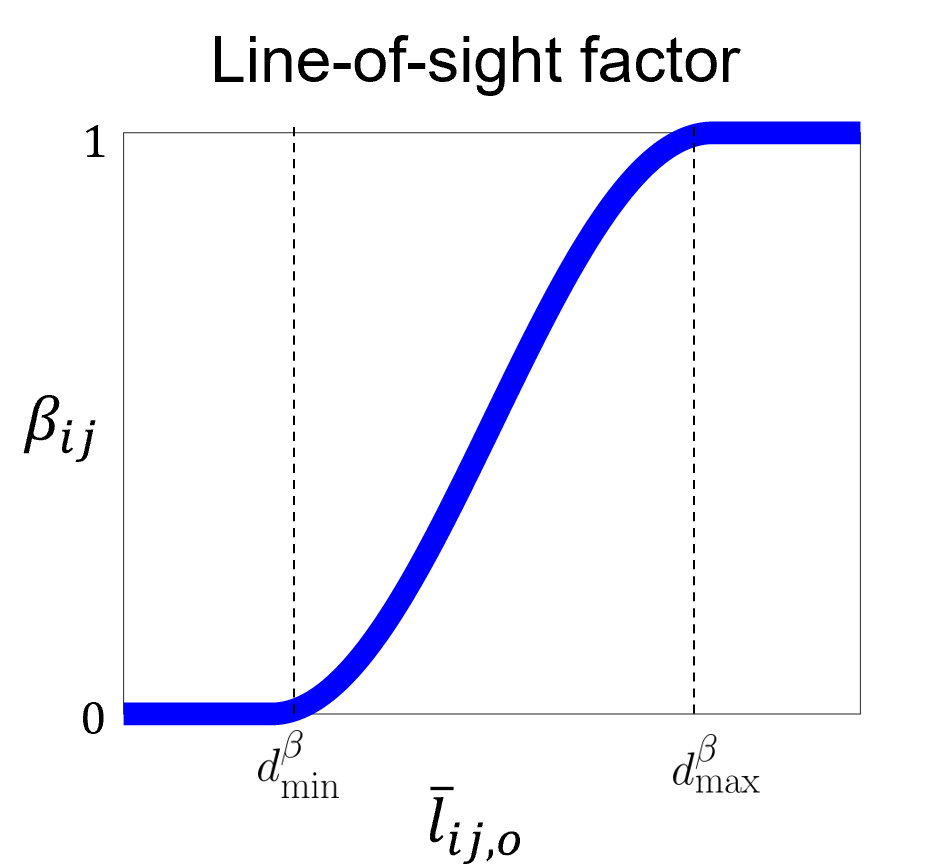}}\hfill
  \subfloat[]{\includegraphics[width=0.25\linewidth]{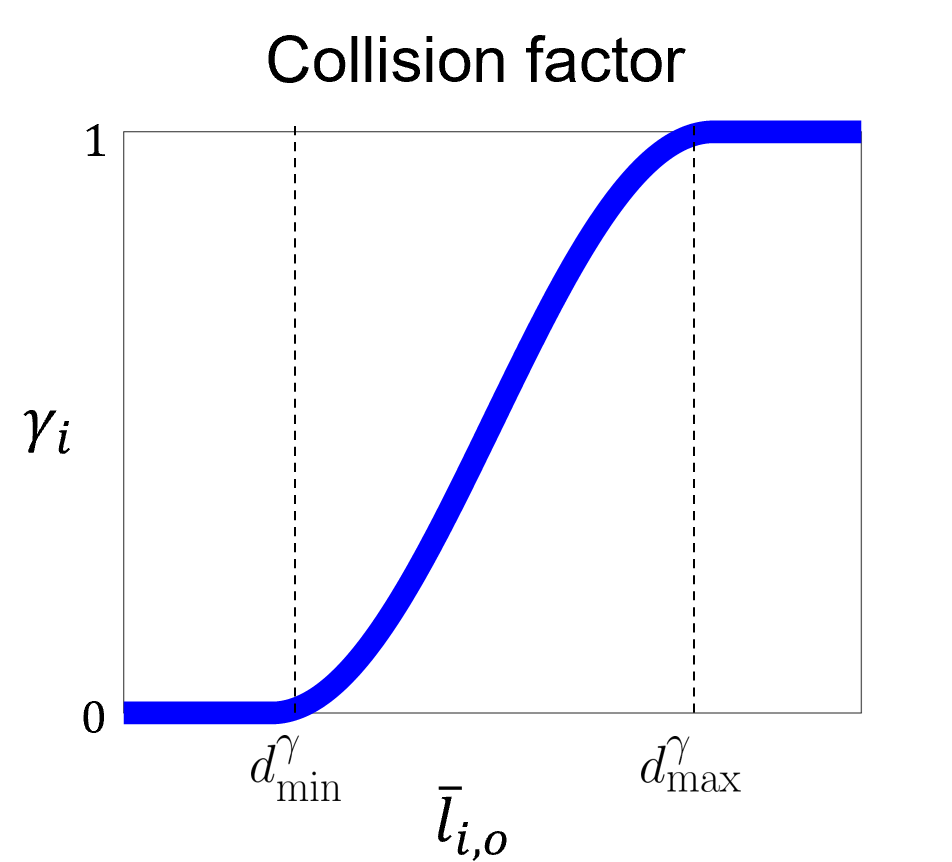}}\hspace{1.5cm}\hfill
  \caption{Factors that define the edge weights of our weighted graph using Equation~(\ref{eqn:edge-weight-definition}). (a) Communication range factor (Equation~(\ref{eqn:comm-range-factor})), (b) line-of-sight factor (Equation~(\ref{eqn:LOS-factor})), and (c) collision factor (Equation~(\ref{eqn:collision-factor})).}
  \label{fig:edge-weight-factors}
\end{figure}

\subsection{Weighted graph definition}
\label{subsec:edge-weights}
For the edge weights to represent connectivity between robots, it needs to consider the constraints listed in Section~\ref{subsec:connectivity-maintenance}. Thus for our weighted graph, we define the edge weight between any two robots $i$ and $j$ as a product of four factors:
\begin{equation}
    \label{eqn:edge-weight-definition}
    \edgeweight_{ij} = \rangefactor_{ij} \LOSfactor_{ij} \collisionfactor_i \collisionfactor_j,
\end{equation}
where $\rangefactor_{ij}$ is a factor for the communication range constraint, $\LOSfactor_{ij}$ is a factor for the line-of-sight constraint, and $\collisionfactor_i$ and $\collisionfactor_j$ are factors for the collision avoidance constraints of the two robots. Note that Equation~(\ref{eqn:edge-weight-definition}) ensures that the edge weight $\edgeweight_{ij}$ is equal to $0$ (i.e., robots $i$ and $j$ are not connected) if any of the factors $\rangefactor_{ij}$, $\LOSfactor_{ij}$, $\collisionfactor_i$ or $\collisionfactor_j$ is equal to $0$ (happens when the corresponding constraint is not satisfied). Next, we define \textit{smooth} functions for each of these factors which allows us to compute gradient-based control inputs as discussed later in Section~\ref{subsec:gradient-controller}.
\begin{enumerate}
    \item Communication range factor $\rangefactor_{ij}$: This factor indicates how well robots $i$ and $j$ are connected in terms of being within a maximum communication range. We begin by defining a conservative measure $\distmeasure_{ij}$ for the range between two robots while accounting for their deviations from their nominal positions up to a desired confidence level. $\distmeasure_{ij}$ is defined as:
    \begin{equation}
        \label{eqn:dist-measure-ij}
        \distmeasure_{ij} = \left \| \check{\state}_i - \check{\state}_j \right \|_2 + s\sqrt{\bar{\eigenvalue}^{\totalstatecov_i}} + s\sqrt{\bar{\eigenvalue}^{\totalstatecov_j}},
    \end{equation}
    where $\check{\state}_i$ and $\check{\state}_j$ are the nominal positions, $s$ is a scalar value obtained using the chi-square distribution for a desired confidence level~\citep{hoover1984algorithms}, and $\bar{\eigenvalue}^{\totalstatecov_i}$ and $\bar{\eigenvalue}^{\totalstatecov_j}$ are the largest eigenvalues of the covariance matrices $\totalstatecov_i$ and $\totalstatecov_j$. Later in Section~\ref{sec:results}, we demonstrate that a $3\sigma$ confidence level (reflecting a ${\small \sim} 99.7\%$ confidence level) results in acceptable connectivity maintenance performance. Fig.~\ref{fig:weighted-graph-lenghts}(a) illustrates the conservative measure $\distmeasure_{ij}$.
    
    Given $\distmeasure_{ij}$ and given a maximum communication range $\commrange$, we define the communication range factor in Equation~(\ref{eqn:edge-weight-definition}) as:
    \begin{equation}
        \label{eqn:comm-range-factor}
        \rangefactor_{ij} = \begin{cases} \ \ \ \ \ \ \ \ \ \ \ 1 & 0 \leq \distmeasure_{ij} \leq \commrange_0 \\
\frac{1}{2} + \frac{1}{2}\cos{ \left [ \frac{\pi(\distmeasure_{ij} - \commrange_0)}{\commrange - \commrange_0 }   \right ] } & \commrange_0 < \distmeasure_{ij} \leq \commrange \\
\ \ \ \ \ \ \ \ \ \ \ 0 & \distmeasure_{ij} > \commrange
\end{cases},
    \end{equation}
    where $\commrange_0$ is a parameter such that $\commrange_0 < \commrange$. Fig.~\ref{fig:edge-weight-factors}(a) shows how $\rangefactor_{ij}$ varies with $\distmeasure_{ij}$.
    
    \item Line-of-sight range factor $\LOSfactor_{ij}$: This factor indicates how well robots $i$ and $j$ are connected in terms of being in line-of-sight of each other. Similar to $\distmeasure_{ij}$ in Equation~(\ref{eqn:dist-measure-ij}) we define a conservative measure $\distmeasure_{ij,o}$ for how close the line-of-sight vector between two robots is to nearby obstacles, as illustrated in Fig.~\ref{fig:weighted-graph-lenghts}(b). Thus, $\distmeasure_{ij,o}$ is defined as: 
    \begin{equation}
        \label{eqn:dist-measure-ij-o}
        \distmeasure_{ij,o} = \left \| {\state}_{l,ij} - {\state}_{\LOSfactor,ij} \right \|_2 - s\sqrt{\max{(\bar{\eigenvalue}^{\totalstatecov_i},\bar{\eigenvalue}^{\totalstatecov_j})}},
    \end{equation}
    where $\state_{\LOSfactor,ij}$ is the closest obstacle point, $s$ is a scalar value reflecting the desired confidence level~\citep{hoover1984algorithms} (similar to Equation~(\ref{eqn:comm-range-factor})), $\bar{\eigenvalue}^{\totalstatecov_i}$ and $\bar{\eigenvalue}^{\totalstatecov_j}$ are the largest eigenvalues of the covariance matrices $\totalstatecov_i$ and $\totalstatecov_j$, and $\state_{l,ij}$ is the point on the line segment connecting $\check{\state}_i$ and $\check{\state}_j$ that is closest to $\state_{\LOSfactor,ij}$, and can be expressed as:
    \begin{equation}
        \label{eqn:state-closest-to-obst}
        \state_{l,ij} = \zeta \check{\state}_i + (1-\zeta) \check{\state}_j,
    \end{equation}
    where $\zeta \in [0,1]$. Note that finding the closest obstacle point $\state_{\LOSfactor,ij}$ can be computationally challenging in practice, especially in the presence of a large number of obstacles. Techniques such as efficient nearest neighbor searches~\citep{atramentov2002efficient} have been explored to reduce this computational load and are beyond the scope of this paper.
    
    Given $\distmeasure_{ij,o}$ and given parameters for the minimum and maximum desired distances ($\losrange_{min}$ and $\losrange_{max}$) from the obstacle, we define the line-of-sight factor in Equation~(\ref{eqn:edge-weight-definition}) as:
    \begin{equation}
        \label{eqn:LOS-factor}
        \LOSfactor_{ij} = \begin{cases} \ \ \ \ \ \ \ \ \ \ \ 1 & \distmeasure_{ij,o} > \losrange_{max} \\
\frac{1}{2} + \frac{1}{2}\cos{ \left [ \frac{\pi(\losrange_{max} - \distmeasure_{ij,o})}{\losrange_{max} - \losrange_{min} }   \right ] } & \losrange_{min} < \distmeasure_{ij,o} \leq \losrange_{max} \\
\ \ \ \ \ \ \ \ \ \ \ 0 & \distmeasure_{ij,o} \leq \losrange_{min}
\end{cases}.
    \end{equation}
    Fig.~\ref{fig:edge-weight-factors}(b) shows how $\LOSfactor_{ij}$ varies with $\distmeasure_{ij,o}$.
    
    \item Collision factor $\collisionfactor_i$: This factor indicates how close robot $i$ is to a collision. Here again we define a conservative measure $\distmeasure_{i,o}$ of the distance from the robot to the nearest possible collision point. This nearest collision point can be a neighboring robot or a nearby obstacle. As mentioned above, techniques to reduce the computational load in obtaining this nearest collision point are beyond the scope of this paper. Fig.\ref{fig:weighted-graph-lenghts}(c) illustrates the conservative measure for two robots $i$ and $j$, i.e., $\distmeasure_{i,o}$ and $\distmeasure_{j,o}$. We define the measure $\distmeasure_{i,o}$ as:
    \begin{equation}
        \label{eqn:dist-measure-l-i-o}
        \distmeasure_{i,o} = \left \| \check{\state}_i - {\state}_{\collisionfactor,i} \right \|_2 - s\sqrt{\bar{\eigenvalue}^{\totalstatecov_i}} - b_{i,o}.
    \end{equation}
    where $\check{\state}_i$ is the nominal position and $s$ is a scalar value reflecting the desired confidence level~\citep{hoover1984algorithms} (similar to Equation~(\ref{eqn:comm-range-factor})). Here if the nearest collision point for robot $i$ is robot $j$, then $\state_{\collisionfactor,i} = \check{\state}_j$ and $b_{i,o} = s\sqrt{\bar{\eigenvalue}^{\totalstatecov_j}}$. Otherwise, if the nearest collision point for robot $i$ is an obstacle then $\state_{\collisionfactor,i}$ is the center of the obstacle and $b_{i,o}$ represents the width of the obstacle. Given $\distmeasure_{i,o}$ and given parameters for the minimum and maximum desired distances ($\collrange_{min}$ and $\collrange_{max}$), we define the collision factor in Equation~(\ref{eqn:edge-weight-definition}) as:
    \begin{equation}
        \label{eqn:collision-factor}
        \collisionfactor_{i} = \begin{cases} \ \ \ \ \ \ \ \ \ \ \ 1 & \distmeasure_{i,o} > \collrange_{max} \\
\frac{1}{2} + \frac{1}{2}\cos{ \left [ \frac{\pi(\collrange_{max} - \distmeasure_{i,o})}{\collrange_{max} - \collrange_{min} }   \right ] } & \collrange_{min} < \distmeasure_{i,o} \leq \collrange_{max} \\
\ \ \ \ \ \ \ \ \ \ \ 0 & \distmeasure_{i,o} \leq \collrange_{min}
\end{cases}.
    \end{equation}
    Fig.~\ref{fig:edge-weight-factors}(c) shows how $\collisionfactor_{i}$ varies with $\distmeasure_{i,o}$. Note that similar to $\collisionfactor_i$, the collision factor $\collisionfactor_j$ indicates how close robot $j$ is to a collision.
\end{enumerate}

Thus, given the nominal positions $\check{\state}_{i}$ and covariance matrices $\totalstatecov_{i}$ for all robots in the system, Equations~(\ref{eqn:edge-weight-definition})-(\ref{eqn:collision-factor}) can be used to obtain the edge weights for our weighted graph. The main objective of these edge weights is to model the connectivity constraints listed in Section~\ref{subsec:connectivity-maintenance}. If desired, alternate functions (such as exponential-based functions presented in~\citep{Sabattini2013} and~\citep{yang2010decentralized}) can be used to model these constraints as long as the edge weights remain in the range $[0,1]$ (as described in Section~\ref{sec:preliminaries}) and are differentiable (required by our gradient-based controller described later in Section~\ref{subsec:gradient-controller}). The gradients of these edge weights with respect to the robot nominal positions are eventually used to compute robot control inputs.

Note that the factors $\rangefactor$ (Equation~(\ref{eqn:comm-range-factor})), $\LOSfactor$ (Equation~(\ref{eqn:LOS-factor})) and $\collisionfactor$ (Equation~(\ref{eqn:collision-factor})) conservatively represent the connectivity constraints listed in Section~\ref{subsec:connectivity-maintenance} since these factors depend on conservative measures defined in Equations~(\ref{eqn:dist-measure-ij}), (\ref{eqn:dist-measure-ij-o}) and (\ref{eqn:dist-measure-l-i-o}). For instance, the conservative measure for the range between two robots (shown in Fig.~\ref{fig:weighted-graph-lenghts}(a)) is typically larger than the true distance (greater than $99.7\%$ of the time for a $3\sigma$ confidence level), consequently leading the communication range factor $\rangefactor$ to have a conservative representation of the communication range constraint. Thus, the algebraic connectivity of our weighted graph ${\eigenvalue}_2$ conservatively measures the true algebraic connectivity of the system $\bar{\eigenvalue}_2$. A detailed analysis comparing ${\eigenvalue}_2$ and $\bar{\eigenvalue}_2$ can be found in our prior work~\citep{shetty2020connectivity}. Hence, in order to maintain $\bar{\eigenvalue}_2 > \AClimit$ according to the connectivity maintenance requirement stated in Section~\ref{sec:problem_formulation}, our DCMU algorithm attempts to maintain ${\eigenvalue}_2 > \AClimit$ as detailed later in Section~\ref{subsec:gradient-controller}.



\subsection{Decentralized power iteration}
\label{subsec:AC-estimation}

In order to compute the nominal control input for connectivity maintenance, each robot needs to obtain information of how well connected the weighted graph is. To obtain this information in a decentralized manner, we implement a decentralized power iteration method similar to previous works~\citep{yang2010decentralized,Sabattini2013}. Here each robot communicates in a decentralized manner, i.e., only with other robots it is connected to, and shares information related to its estimate of the system connectivity. Based on the information received from connected robots, each robot is able to gradually improve its estimates of the system connectivity. Further details of the method can be found in~\citep{Sabattini2013}. At the end of each iteration of the power method each robot $i$ estimates the following two quantities:
\begin{enumerate}
    \item The $i^{th}$ component of the Fiedler vector (Section~\ref{subsec:edge-weights}) of the weighted graph, $\tilde{\eigenvector}_{2,i}^{(i)}$. This contains information of how well robot $i$ is connected to the system.
    \item The algebraic connectivity (Section~\ref{subsec:edge-weights}) of the weighted graph, $\tilde{\eigenvalue}_{2,i}$. This contains information of how well connected the entire multi-robot system is.
\end{enumerate}
For additional details of the decentralized power iteration method, we refer our readers to~\citep{Sabattini2013}.

\subsection{Decentralized gradient-based controller}
\label{subsec:gradient-controller}

As mentioned in Section~\ref{subsec:connectivity-maintenance}, the objective of the controller is to derive the nominal control input $\check{\control}_{i,t}$ for each follower robot $i$ to maintain the algebraic connectivity $\bar{\eigenvalue}_2$ above a specified lower limit $\AClimit$. In order to achieve this, we instead maintain the algebraic connectivity of our weighted graph $\AC$ above $\AClimit$, which is a conservative representation of the multi-robot system connectivity as explained at the end of Section~\ref{subsec:edge-weights}.

\begin{figure}[t]
  \centering
  \hspace{3cm} \subfloat[]{\includegraphics[width=0.25\linewidth]{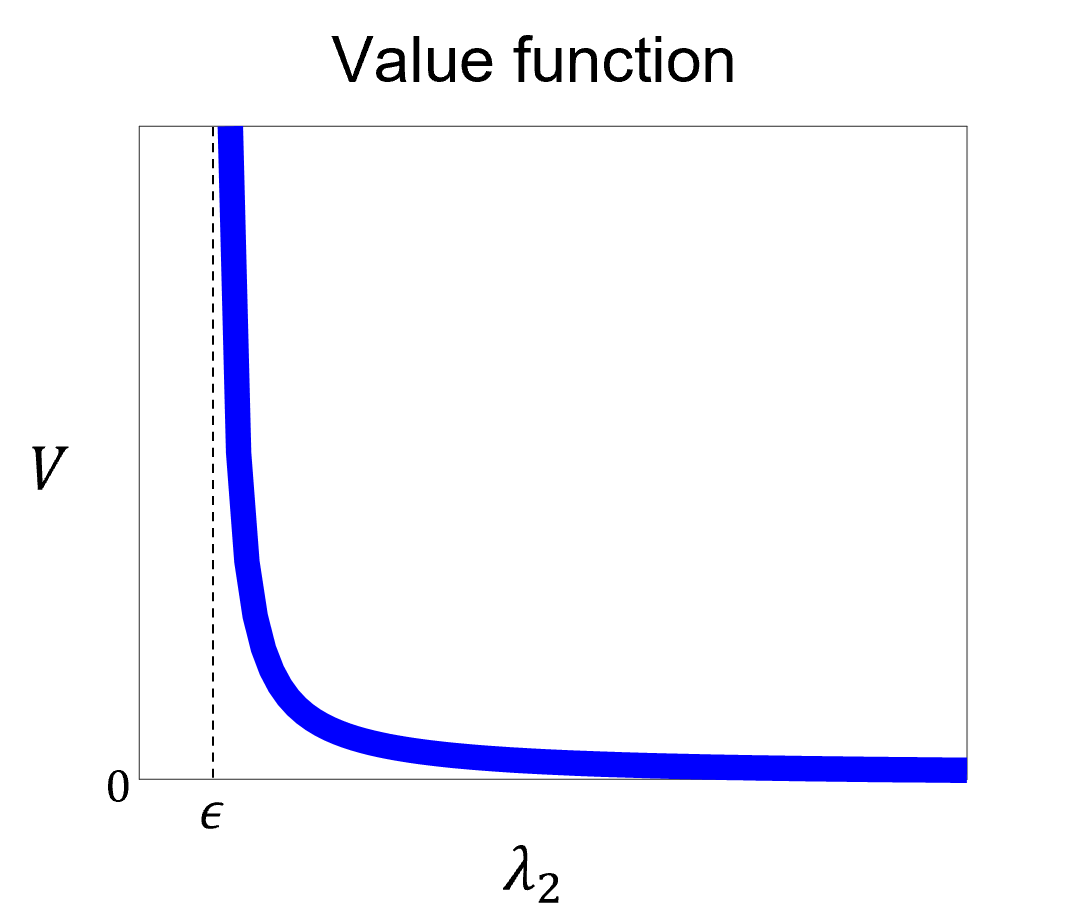}}\hfill
  \subfloat[]{\includegraphics[width=0.25\linewidth]{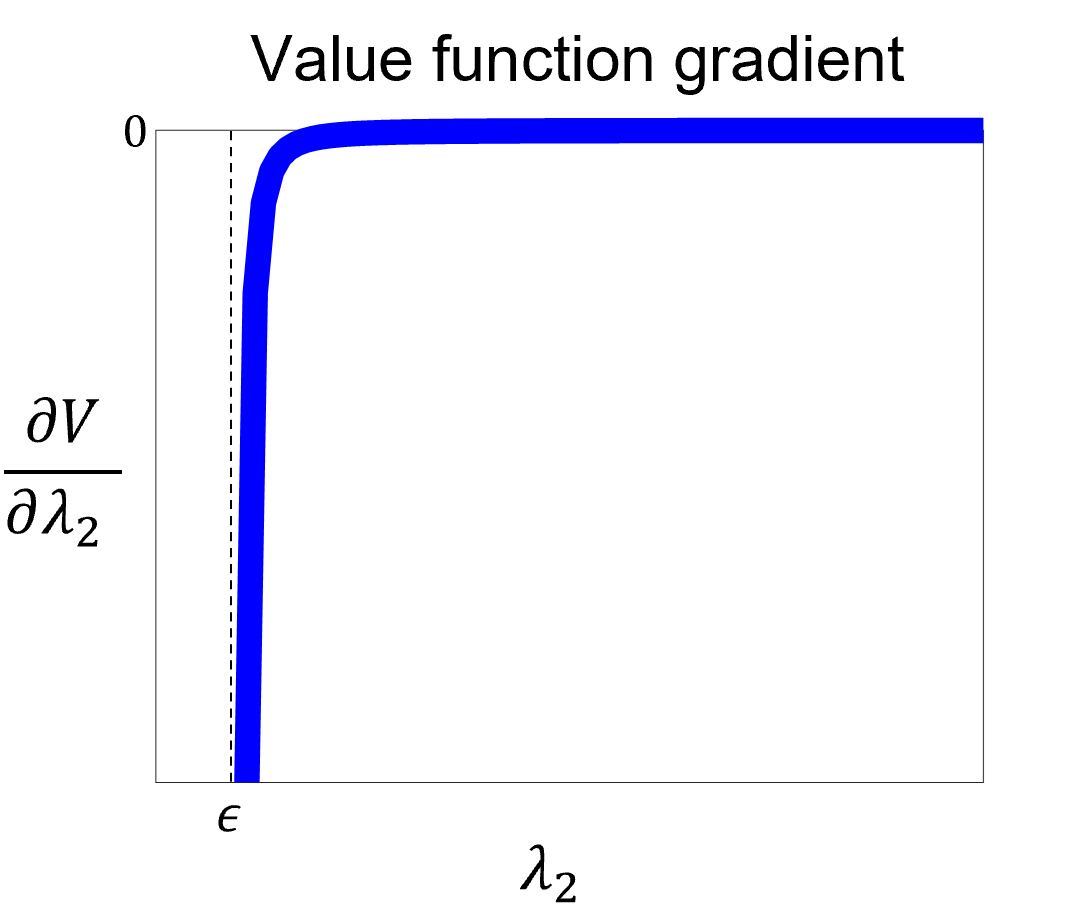}}\hspace{3cm}\hfill
  \caption{(a) The value function~\citep{Sabattini2013} defined in Equation~(\ref{eqn:value-function}), along with (b) the gradient of the value function required to compute the nominal control input in Equation~(\ref{eqn:nominal-input}).}
  \label{fig:value-function}
\end{figure}

We use a similar approach as~\citep{Sabattini2013}, where we first define a value function $\valuefunc$ that depends on algebraic connectivity of our weighted graph $\AC$, and then design a gradient-based controller for the follower robots to increase $\AC$. The value function is defined as:
\begin{equation}
    \label{eqn:value-function}
    \valuefunc (\AC) = \begin{cases} \coth{( \AC - \AClimit )} & \AC > \AClimit \\
\ \ \ \ \ \ \ \ \ \ 0 & \text{otherwise}
\end{cases}.
\end{equation}
Fig.~\ref{fig:value-function}(a) shows the variation of the value function with $\AC$. To maintain connectivity in the system, we seek to adjust the robots' nominal positions in order to perform a gradient descent of the value function. This consequently results in increasing the value of $\AC$ as seen in Fig.~\ref{fig:value-function}(a). Thus, after simplification of the value function gradient~\citep{yang2010decentralized,Sabattini2013}, the desired change in the robot's nominal position can be expressed as:
\begin{equation}
    \label{eqn:nominal-position-change}
    \Delta \check{\state}_i = \left ( -\gradient{\valuefunc(\AC)}{\check{\state}_i} \right ) = \left ( -\gradient{\valuefunc(\AC)}{\AC} \right ) \gradient{\AC}{\check{\state}_i} = \left (- \gradient{\valuefunc(\AC)}{\AC} \right ) \sum_{j=1}^{\N} \gradient{\edgeweight_{ij}}{\check{\state}_i} \left ( \tilde{\eigenvector}_{2,i}^{(i)} - \tilde{\eigenvector}_{2,j}^{(j)} \right )^2,
\end{equation}
where the gradient of the value function $\gradient{\valuefunc(\AC)}{\AC}$ is evaluated at $\tilde{\eigenvalue}_{2,i}$. Here $\tilde{\eigenvalue}_{2,i}$, $\tilde{\eigenvector}_{2,i}^{(i)}$ and $\tilde{\eigenvector}_{2,j}^{(j)}$ are obtained from the decentralized power iteration method as explained in Section~\ref{subsec:AC-estimation}. Fig.~\ref{fig:value-function}(b) shows the gradient of the value function. Note that as $\AC$ approaches $\AClimit$, the magnitude of the gradient sharply increases resulting in a large $\Delta \check{\state}_i$. This consequently leads to the follower robots moving faster (limited by their maximum velocity) to maintain connectivity when $\AC$ approaches $\AClimit$.

Note that if desired alternate value functions can be used instead of Equation~(\ref{eqn:value-function}). Given that the gradient of the value function affects the change in the robots' nominal positions (as shown in Equation~(\ref{eqn:nominal-input})),~\citep{Sabattini2013} mentions the required properties for this function. It needs to be continuously differentiable, non-negative and non-increasing $\forall \ \AC > \AClimit$. Additionally, it needs to suddenly increase as $\AC$ approaches $\AClimit$ (resulting in large changes in positions when the system is near loss of connectivity) and it needs to approach a constant value for large values of $\AC$ (resulting in small changes in positions when the system is \textit{well} connected). 

Next, using Equations~(\ref{eqn:nominal-state}) and (\ref{eqn:nominal-position-change}), the nominal control input can be obtained as:
\begin{equation}
    \label{eqn:nominal-input}
    \check{\control}_{i} = \controlmatrix^{-1} \Delta \check{\state}_i = \frac{1}{\Dt} \left ( -\gradient{\valuefunc(\AC)}{\AC} \right ) \sum_{j=1}^{\N} \gradient{\edgeweight_{ij}}{\check{\state}_i} \left ( \tilde{\eigenvector}_{2,i}^{(i)} - \tilde{\eigenvector}_{2,j}^{(j)} \right )^2,
\end{equation}
where the control input matrix $\controlmatrix = (\Dt)\eye$ as defined in Equation~(\ref{eqn:motion-model}). In order to compute the nominal control input $\check{\control}_{i}$ using the above equation, we require the gradients of the edge weights $\edgeweight_{ij}$ with respect to the robot nominal position $\check{\state}_i$, i.e., $\gradient{\edgeweight_{ij}}{\check{\state}_i}$. Given the definition of the edge weight in Equation~(\ref{eqn:edge-weight-definition}), the required gradient can be expressed as:
\begin{equation}
    \label{eqn:edge-weight-gradient}
    \gradient{\edgeweight_{ij}}{\check{\state}_i} = \gradient{\rangefactor_{ij}}{\check{\state}_i} \LOSfactor_{ij} \collisionfactor_i \collisionfactor_j + \rangefactor_{ij} \gradient{\LOSfactor_{ij}}{\check{\state}_i} \collisionfactor_i \collisionfactor_j + \rangefactor_{ij} \LOSfactor_{ij} \gradient{\collisionfactor_i}{\check{\state}_i} \collisionfactor_j + \rangefactor_{ij} \LOSfactor_{ij} \collisionfactor_i \gradient{\collisionfactor_j}{\check{\state}_i}.
\end{equation}
In the remainder of this subsection we proceed to derive the expressions for the gradient of the factors $\rangefactor_{ij}$, $\LOSfactor_{ij}$, $\collisionfactor_i$ and $\collisionfactor_j$ with respect to $\check{\state}_i$. From Equation~(\ref{eqn:comm-range-factor}), the gradient of $\rangefactor_{ij}$ can be expressed as:
\begin{equation}
    \label{eqn:rangefactor-gradient}
    \gradient{\rangefactor_{ij}}{\check{\state}_i} = -\frac{\pi}{2(\commrange - \commrange_0)} \sin \left [ \frac{\pi (\distmeasure_{ij} - \commrange_0)}{\commrange - \commrange_0} \right ] \gradient{\distmeasure_{ij}}{\check{\state}_i},
\end{equation}
where using Equation~(\ref{eqn:dist-measure-ij}), the gradient of $\distmeasure_{ij}$ with respect to the $m^{th}$ element of $\check{\state}_i$ is:
\begin{equation}
    \label{eqn:gradient-distmeasure-ij}
    \gradient{\distmeasure_{ij}}{\check{\state}_i^{(m)}} = \frac{ \check{\state}_i^{(m)} - \check{\state}_j^{(m)}}{ \left \| \check{\state}_i - \check{\state}_j \right \|_2 } + \frac{s}{2 \sqrt{\bar{\eigenvalue}^{\totalstatecov_i}}} \gradient{\bar{\eigenvalue}^{\totalstatecov_i}}{\check{\state}_i^{(m)}}.
\end{equation}
Next, for the gradient of $\LOSfactor_{ij}$, using Equation~(\ref{eqn:LOS-factor}) we get:
\begin{equation}
    \label{eqn:LOSfactor-gradient}
    \gradient{\LOSfactor_{ij}}{\check{\state}_i} = \frac{\pi}{2 (\losrange_{max} - \losrange_{min})} \sin \left [ \frac{\pi (\losrange_{max} - \distmeasure_{ij,o})}{\losrange_{max} - \losrange_{min}} \right ] \gradient{\distmeasure_{ij,o}}{\check{\state}_i},
\end{equation}
where using Equation~(\ref{eqn:dist-measure-ij-o}) the gradient of $\distmeasure_{ij,o}$ with respect to the $m^{th}$ element of $\check{\state}_i$ is:
\begin{equation}
    \label{eqn:gradient-distmeasure-ij-o}
    \gradient{\distmeasure_{ij,o}}{\check{\state}_i^{(m)}} = \zeta \frac{ (\check{\state}_l^{(m)} - \state_o^{(m)}) }{ \left \| \check{\state}_l - {\state}_o \right \|_2 } - \frac{s}{2 \sqrt{\max{(\bar{\eigenvalue}^{\totalstatecov_i},\bar{\eigenvalue}^{\totalstatecov_j})}}} \gradient{\max({\bar{\eigenvalue}^{\totalstatecov_i},\bar{\eigenvalue}^{\totalstatecov_j}})}{\check{\state}_i^{(m)}}.
\end{equation}
Using Equation~(\ref{eqn:collision-factor}), the gradient of $\collisionfactor_i$ can be expressed as:
\begin{equation}
    \label{eqn:collisionfactor-i-gradient}
    \gradient{\collisionfactor_i}{\check{\state}_i} = \frac{\pi}{2 (\collrange_{max} - \collrange_{min})} \sin \left [ \frac{\pi (\collrange_{max} - \distmeasure_{i,o})}{\collrange_{max} - \collrange_{min}} \right ] \gradient{\distmeasure_{i,o}}{\check{\state}_i},
\end{equation}
where using Equation~(\ref{eqn:dist-measure-l-i-o}) the gradient of $\distmeasure_{i,o}$ with respect to the $m^{th}$ element of $\check{\state}_i$ is:
\begin{equation}
    \label{eqn:gradient-distmeasure-i-o}
    \gradient{\distmeasure_{i,o}}{\check{\state}_i^{(m)}} = \frac{ (\check{\state}_i^{(m)} - \state_o^{(m)}) }{ \left \| \check{\state}_i - {\state}_o \right \|_2 } - \frac{s}{2 \sqrt{\bar{\eigenvalue}^{\totalstatecov_i}}} \gradient{\bar{\eigenvalue}^{\totalstatecov_i}}{\check{\state}_i^{(m)}}.
\end{equation}
Since $\collisionfactor_j$ indicates how close robot $j$ is to a collision, its gradient with respect to $\check{\state}_i$ is non-zero only when the closest collision point for robot $j$ is robot $i$. As explained below Equation~(\ref{eqn:dist-measure-l-i-o}), this implies $\state_{\collisionfactor,j} = \check{\state}_i$ and $b_{j,o} = s\sqrt{\bar{\eigenvalue}^{\totalstatecov_i}}$. Thus, if the closest collision point for robot $j$ is robot $i$, then using Equation~(\ref{eqn:collision-factor}) the gradient of $\collisionfactor_j$ can be expressed as:
\begin{equation}
    \label{eqn:collisionfactor-j-gradient}
    \gradient{\collisionfactor_j}{\check{\state}_i} = \frac{\pi}{2 (\collrange_{max} - \collrange_{min})} \sin \left [ \frac{\pi (\collrange_{max} - \distmeasure_{j,o})}{\collrange_{max} - \collrange_{min}} \right ] \gradient{\distmeasure_{j,o}}{\check{\state}_i},
\end{equation}
where using Equation~(\ref{eqn:dist-measure-l-i-o}) the gradient of $\distmeasure_{j,o}$ with respect to the $m^{th}$ element of $\check{\state}_i$ is:
\begin{equation}
    \label{eqn:gradient-distmeasure-j-o}
    \gradient{\distmeasure_{j,o}}{\check{\state}_i^{(m)}} = -\frac{ (\check{\state}_j^{(m)} - \check{\state}_i^{(m)}) }{ \left \| \check{\state}_j - \check{\state}_i \right \|_2 } - \frac{s}{2 \sqrt{\bar{\eigenvalue}^{\totalstatecov_i}}} \gradient{\bar{\eigenvalue}^{\totalstatecov_i}}{\check{\state}_i^{(m)}}.
\end{equation}

In order to compute the gradients of the factors using Equations~(\ref{eqn:rangefactor-gradient}), (\ref{eqn:LOSfactor-gradient}), (\ref{eqn:collisionfactor-i-gradient}) and (\ref{eqn:collisionfactor-j-gradient}), we need to derive the gradients of the eigenvalues of covariance matrices $\totalstatecov_i$ and $\totalstatecov_j$ required in Equations~(\ref{eqn:gradient-distmeasure-ij}), (\ref{eqn:gradient-distmeasure-ij-o}), (\ref{eqn:gradient-distmeasure-i-o}) and (\ref{eqn:gradient-distmeasure-j-o}). The equations in Section~\ref{subsec:robot-description} show how the covariance matrix $\totalstatecov_i$ for a robot varies with its position $\check{\state}_i$. From Equation~(\ref{eqn:total-state-cov-dist}), note that $\totalstatecov_i = \statecov_i + \stateestcov_i$. Thus, as the robot nominal position $\check{\state}_i$ changes from time instant $t$ to $t+1$, the covariance matrix updates to $\totalstatecov_{i,t+1}$ as:
\begin{align} 
\label{eqn:total-covariance-matrix-update}
\totalstatecov_{i,t+1} &= \statecov_{i,t+1} + \stateestcov_{i,t+1} \\ 
{} &=  \bar{\statecov}_{i,t+1} - \kalmangain_{i,t+1} \bar{\statecov}_{i,t+1} + (\eye - \controlmatrix \feedbackgain_{i,t+1}) \stateestcov_{i,t} (\eye - \controlmatrix \feedbackgain_{i,t+1})^\top + \kalmangain_{i,t+1} \bar{\statecov}_{i,t+1} \\
{}& = \statecov_{i,t} + \motioncov_{i,t+1} + (\eye - \controlmatrix \feedbackgain_{i,t+1}) \stateestcov_{i,t} (\eye - \controlmatrix \feedbackgain_{i,t+1})^\top.
\end{align}
Note that the updated covariance matrix $\totalstatecov_{i,t+1}$ only depends on $\statecov_{i,t}$, $\motioncov_{i,t+1}$, $\controlmatrix$, $\feedbackgain_{i,t+1}$ and $\stateestcov_{i,t}$, and does not depend on the new nominal position $\check{\state}_{i,t+1}$. Consequently, the eigenvalue $\bar{\eigenvalue}^{\totalstatecov_i}$ also does not depend on $\check{\state}_i$, which implies $\gradient{\bar{\eigenvalue}^{\totalstatecov_i}}{\check{\state}_i^{(m)}} = 0$. Thus, the terms involving gradients of the eigenvalues in Equations~(\ref{eqn:gradient-distmeasure-ij}),~(\ref{eqn:gradient-distmeasure-ij-o}),~(\ref{eqn:gradient-distmeasure-i-o}) and~(\ref{eqn:gradient-distmeasure-j-o}) are equal to zero.

In summary, our DCMU algorithm obtains nominal control inputs $\check{\control}_i$ for the follower robots using Equation~(\ref{eqn:nominal-input}), which relies on quantities estimated using the decentralized power iteration method mentioned in Section~\ref{subsec:AC-estimation} and relies on gradients of our weighted graph edge weights derived in Equations~(\ref{eqn:edge-weight-gradient})-(\ref{eqn:total-covariance-matrix-update}). In Equation~(\ref{eqn:nominal-input}) note that the magnitude of the nominal control input $\check{\control}_i$ depends directly on the magnitude of the value function gradient (shown in Fig.~\ref{fig:value-function}(b)), which grows large as the system algebraic connectivity decreases. Thus, when the system is close to losing connectivity, Equation~(\ref{eqn:nominal-input}) obtains large nominal control inputs $\check{\control}_i$ for the follower robots to maintain connectivity.

Note that the connectivity maintenance performance of our DCMU algorithm depends on the nominal trajectories of the leader robots, which are assumed to be obtained from a high-level planner as mentioned in Section~\ref{subsec:connectivity-maintenance}. While our algorithm attempts to move the follower robots such that connectivity is maintained, in some cases the nominal trajectories of the leader robots might be such that connectivity maintenance is not possible; for example, a system with two leader robots and one follower robot where the leader robots move indefinitely in opposite directions. On the other hand, our algorithm does guarantee collision avoidance for the robots with the desired confidence level (greater than $99.7\%$ for $3\sigma$ confidence level). If a robot $i$ gets close to a collision point, the corresponding collision factor $\collisionfactor_i$ decreases, consequently decreasing the weighted graph algebraic connectivity. The nominal control input $\check{\control}_i$ obtained using Equation~(\ref{eqn:nominal-input}) then attempts to move the robot away from the collision point. We validate the connectivity maintenance and collision avoidance performance of our DCMU algorithm on a few multi-robot systems below in Section~\ref{sec:results}.





\section{Simulation Results}\label{sec:results}



In this section we validate the performance of our DCMU algorithm in two 2-dimensional simulation setups: MATLAB and AirSim~\citep{shah2018airsim}. We first present the MATLAB simulations which allow us to quantitatively compare our algorithm with related previous work in~\citep{Sabattini2013}, which we refer to as the Sabattini algorithm for simplicity. We then discuss the AirSim simulations which validate our algorithm on a high-fidelity simulator where the motion of the robots is more realistically simulated. For both simulation setups we first demonstrate our DCMU algorithm on a simple two robot setup with one leader robot and one follower robot, and then proceed to more complex multi-robot configurations. As mentioned in Section~\ref{subsec:connectivity-maintenance}, the leader robots are given predefined nominal trajectories (which we assume are available from some high-level planner such as an exploration strategy), and the follower robots implement nominal control inputs from our DCMU algorithm, i.e., from Equation~(\ref{eqn:nominal-input}). A video containing the simulation results can be viewed online at \url{https://youtu.be/SbE-ejQ_zm8}.

For both our simulation setups we set the discrete time-step $\Dt = \SI{0.2}{\second}$. We assume an inter-robot communication rate of $1000$Hz for the decentralized power iteration method. The motion and sensing model covariance matrices are set as $Q_{i,t} = (\SI{0.02}{\square\meter}) \eye$ and $R_{i,t} = (\SI{5}{\square\meter}) \eye$ respectively. These covariance matrices are used to add zero-mean Gaussian-distributed errors to the state and measurement vectors (Equations~(\ref{eqn:motion-model}) and~(\ref{eqn:sensing-model})) for both the leader and the follower robots in the system. The initial state estimation covariance matrix is set as $P_{i,t} = (\SI{0.1}{\square\meter})\eye$ and the control feedback gain used in Equation~(\ref{eqn:total-control}) is set as $\check{\feedbackgain}_i = (0.14) \eye$. For a $3\sigma$ confidence level (reflecting a ${\small \sim} 99.7\%$ confidence level) in our weighted graph, we obtain the scalar value $s = 3.494$ using the chi-square distribution~\citep{hoover1984algorithms}. For the MATLAB simulations we limit the maximum velocity control input (in each direction) in the motion model (Equation~(\ref{eqn:motion-model})) to be $\SI{2}{\meter\per\second}$, in order to reflect physical constraints of practical robots. The AirSim simulator inherently implements maximum velocity constraints while simulating the robot motion. \new{We} specify the lower limit for the algebraic connectivity of the system to be $\AClimit = 0.01$, where the connectivity maintenance objective is to maintain $\bar{\eigenvalue}_2 > \AClimit$.

\new{For the weighted graph in Section~\ref{subsec:edge-weights}, the parameters $\commrange$, $\losrange_{min}$ and $\collrange_{min}$ represent the maximum desired communication range between robots, the minimum desired distance between robot line-of-sight vectors and obstacles, and the minimum desired distance between robots and their nearest collision points respectively. The parameters $\commrange_0$, $\losrange_{max}$ and $\collrange_{max}$ represent the distances at which the communication range factor, the line-of-sight factor and the collision factors respectively start affecting the nominal control input. For instance, the communication range factor between two robots affects their nominal control inputs only when the conservative measure in Equation~(\ref{eqn:dist-measure-ij}) is greater than $\commrange_0$. Thus, setting the value of $\commrange_0$ close to $\commrange$ results in the communication range factor affecting the nominal control inputs only when the conservative measure (Equation~(\ref{eqn:dist-measure-ij})) is close to $\commrange$, versus setting $\commrange_0$ much smaller than $\commrange$ would result in the nominal control inputs being affected much earlier. In practice, these parameters can be set based on the desired performance of the multi-robot system. For our simulations, we heuristically set these parameter values as: $\commrange = \SI{20}{\meter}$, $\commrange_0 = \SI{18}{\meter}$, $\losrange_{max} = \SI{3}{m}$, $\losrange_{min} = \SI{1}{m}$, $\collrange_{max} = \SI{3}{m}$ and $\collrange_{min} = \SI{1}{m}$.}

\subsection{MATLAB simulations}
\label{subsec:matlab-sims}

For the MATLAB simulations, our primary objective is to compare the connectivity maintenance performance of our algorithm with the Sabattini algorithm~\citep{Sabattini2013} for various multi-robot system configurations. As mentioned in the beginning of Section~\ref{sec:dcm_algorithm}, previous works~\citep{yang2010decentralized,Sabattini2013,robuffo2013passivity,gasparri2017bounded,siligardi2019robust} in essence assume that the robots are at their desired nominal positions. In contrast, our DCMU algorithm accounts for the deviations arising due to motion and sensing uncertainties.

\begin{figure}[b!]
  \centering
  \subfloat{\includegraphics[width=0.8\linewidth]{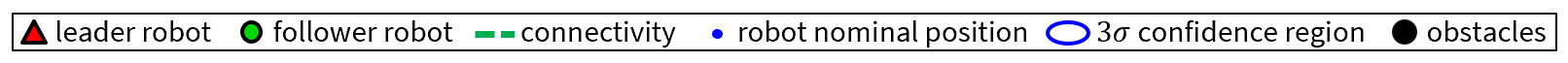}}\addtocounter{subfigure}{-1}\hfill \\
  \subfloat[]{\includegraphics[width=0.24\linewidth]{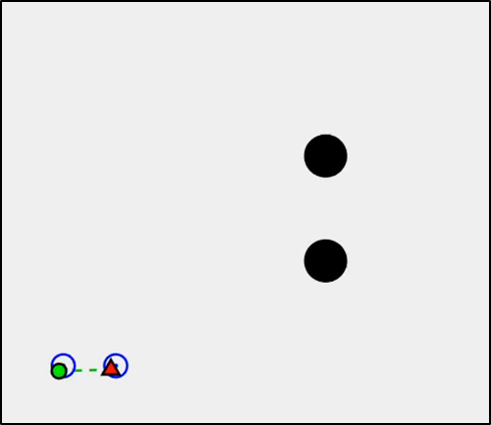}}\hfill
  \subfloat[]{\includegraphics[width=0.24\linewidth]{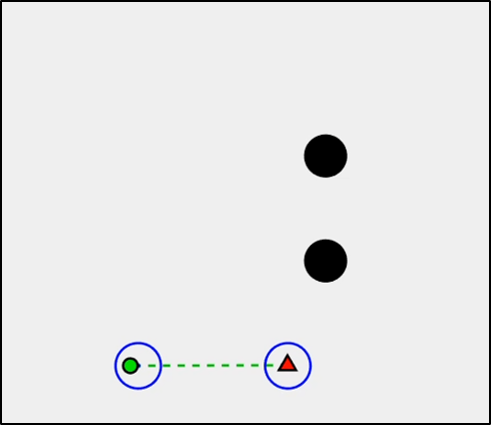}}\hfill
  \subfloat[]{\includegraphics[width=0.24\linewidth]{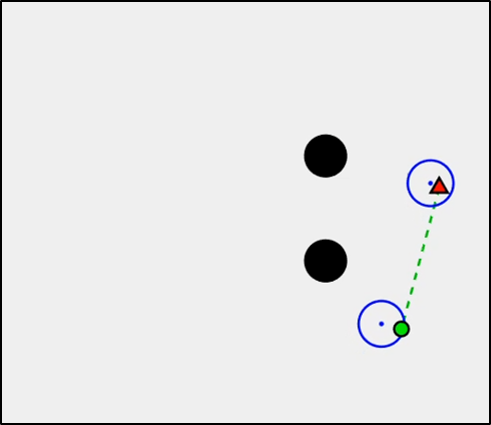}}\hfill
  \subfloat[]{\includegraphics[width=0.24\linewidth]{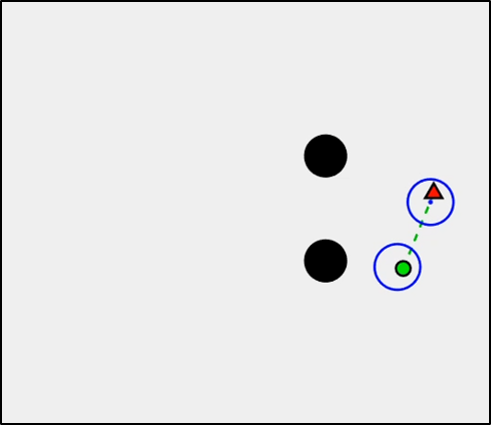}}\hfill \\ \vspace{0.1cm}
  \subfloat[]{\hspace{1cm}\includegraphics[width=0.24\linewidth]{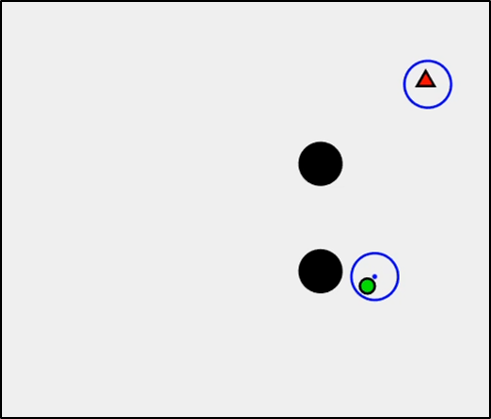}}\hfill
  \subfloat[]{\includegraphics[width=0.24\linewidth]{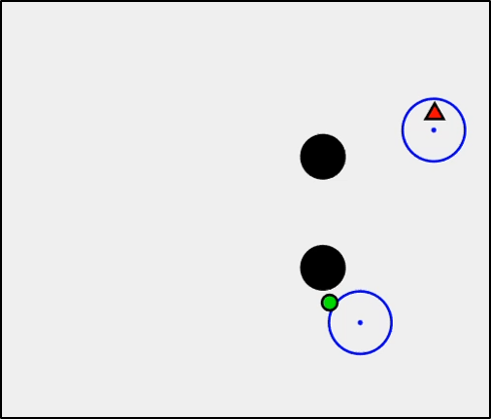}}\hfill
  \subfloat[]{\includegraphics[width=0.24\linewidth]{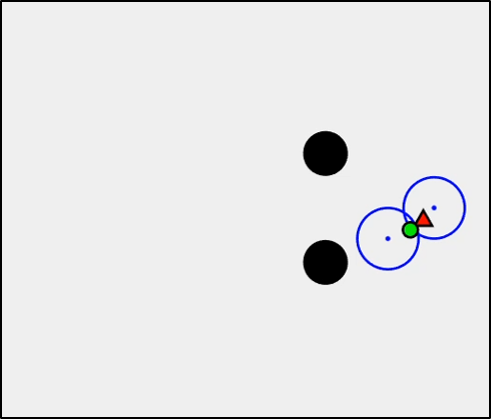}\hspace{1cm}}\hfill
  \caption{MATLAB simulation for a two-robot system. (a)-(d) Snapshots of a simulation run where the follower robot implements our DCMU algorithm. (e)-(g) Snapshots of different simulation runs where connectivity is lost when the follower robot implements the Sabattini algorithm~\citep{Sabattini2013}. The simulation video can be viewed online at \url{https://youtu.be/SbE-ejQ_zm8}.}
  \label{fig:matlab-two-robot-snapshots}
\end{figure}

As shown in Fig.~\ref{fig:matlab-two-robot-snapshots}, we first analyze a simple two robot setup with one leader robot and one follower. The leader robot is provided a nominal trajectory that covers a distance of $\SI{120}{\meter}$ in a duration of $\SI{120}{\second}$. Fig.~\ref{fig:matlab-two-robot-snapshots}(a)-(d) show snapshots of a single simulation run where the follower robot implements our DCMU algorithm to maintain connectivity with the leader robot. Starting from an initial position (Fig.~\ref{fig:matlab-two-robot-snapshots}(a)), the follower robot follows the leader robot to stay within the communication range (Fig.~\ref{fig:matlab-two-robot-snapshots}(b)). As the leader robot moves upwards, the follower robot maintains line-of-sight connectivity while maintaining a safe distance from the obstacles (Fig.~\ref{fig:matlab-two-robot-snapshots}(c)), while also maintaining a safe distance from the leader robot itself (Fig.~\ref{fig:matlab-two-robot-snapshots}(d)). Fig.~\ref{fig:matlab-two-robot-snapshots}(e)-(g) show snapshots of different simulation runs where the follower robot implements the Sabattini algorithm, i.e, does not account for the motion and sensing uncertainties~\citep{Sabattini2013}. We observe that not accounting for these uncertainties can potentially lead to a loss of connection due to the robots getting further away than the communication range (Fig.~\ref{fig:matlab-two-robot-snapshots}(e)), or due to the robots losing line-of-sight communication (Fig.~\ref{fig:matlab-two-robot-snapshots}(f)), or due to a collision (Fig.~\ref{fig:matlab-two-robot-snapshots}(g)).

\begin{figure}[t!]
  \centering
  \includegraphics[width=0.6\linewidth]{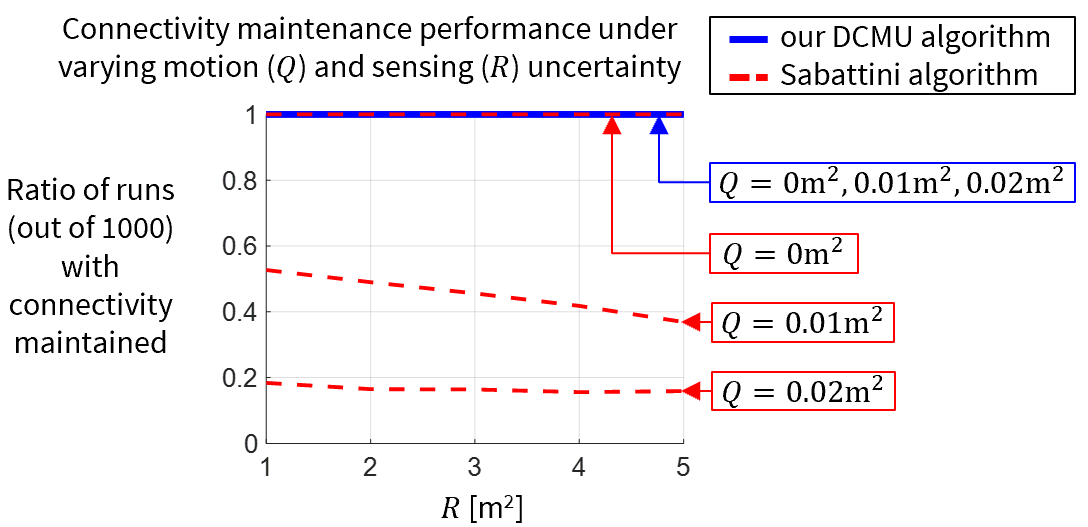}\hfill
  \caption{Quantitative comparison for the two-robot MATLAB simulation shown in Fig~\ref{fig:matlab-two-robot-snapshots}. Our DCMU algorithm performs better than the Sabattini~\citep{Sabattini2013} algorithm under increasing motion and sensing uncertainties.}
  \label{fig:matlab-two-robot-quantitative}
\end{figure}

In Fig.~\ref{fig:matlab-two-robot-quantitative}, we quantitatively compare the connectivity maintenance performance for the above two robot setup with the Sabattini algorithm~\citep{Sabattini2013}. Here we vary the amount of motion and sensing uncertainties in the multi-robot system and perform $1000$ simulation runs for each scenario for both our DCMU algorithm and the Sabattini algorithm. For both these algorithms we compare the ratio of runs for which the algebraic connectivity based on the actual robot positions $\bar{\eigenvalue}_2$ was maintained above the specified lower limit $\AClimit=0.01$ throughout the run, as stated in the problem formulation in Section~\ref{sec:problem_formulation}. Note that for a given set of nominal trajectories for the leader robots, the nominal trajectories for the follower robots obtained using our DCMU algorithm remains the same. However, the actual robot trajectories vary across the $1000$ simulation runs due to the presence of Gaussian-distributed motion and sensing uncertainties. We begin with the trivial case of zero motion model uncertainty, i.e., $Q_{i,t} = (\SI{0}{\square\meter}) \eye$ where the robots exactly follow the nominal trajectories. Here we observe that our DCMU algorithm performs the same as the Sabattini algorithm and maintains $\bar{\eigenvalue}_2 > \AClimit$ throughout all $1000$ simulation runs. We then compare the connectivity maintenance performance for the following motion and sensing model covariance matrices: $Q_{i,t} = (\SI{0.01}{\square\meter}) \eye$ and $Q_{i,t} = (\SI{0.02}{\square\meter}) \eye$; $R_{i,t} = (\SI{1}{\square\meter}) \eye$, $R_{i,t} = (\SI{2}{\square\meter}) \eye$, $R_{i,t} = (\SI{3}{\square\meter}) \eye$, $R_{i,t} = (\SI{4}{\square\meter}) \eye$ and $R_{i,t} = (\SI{5}{\square\meter}) \eye$. Our DCMU algorithm maintains $\bar{\eigenvalue}_2 > \AClimit$ throughout all $1000$ simulation runs for the different motion and sensing uncertainties. However, as the amount of uncertainty is increased, we observe that the connectivity maintenance performance of the Sabattini algorithm decreases sharply.

\begin{figure}[t!]
  \centering
  \subfloat{\hspace{-0.5cm} \includegraphics[width=0.8\linewidth]{images/matlab_legend.png}}\addtocounter{subfigure}{-1}\hfill \\ \hspace{1cm}
  \subfloat[]{\includegraphics[width=0.23\linewidth]{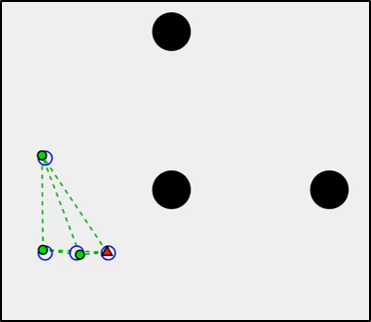}}\hfill
  \subfloat[]{\includegraphics[width=0.23\linewidth]{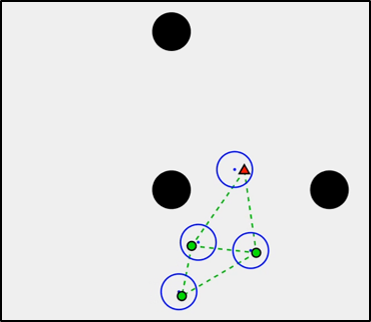}}\hfill
  \subfloat[]{\includegraphics[width=0.23\linewidth]{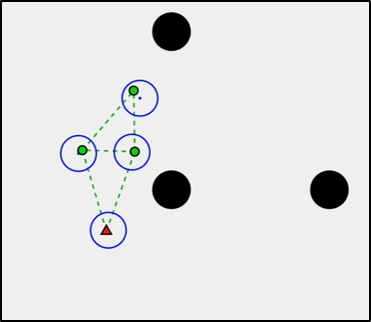}}\hfill \hspace{2cm} \\ \hspace{1cm}
  \subfloat[]{\includegraphics[width=0.23\linewidth]{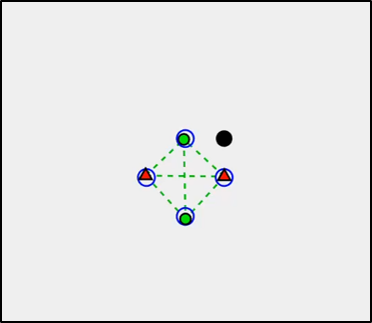}}\hfill
  \subfloat[]{\includegraphics[width=0.23\linewidth]{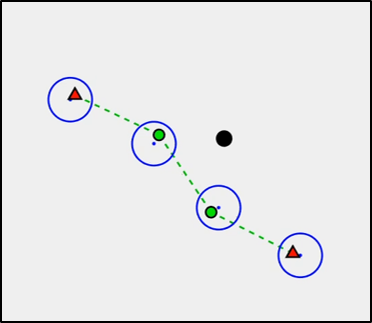}}\hfill
  \subfloat[]{\includegraphics[width=0.23\linewidth]{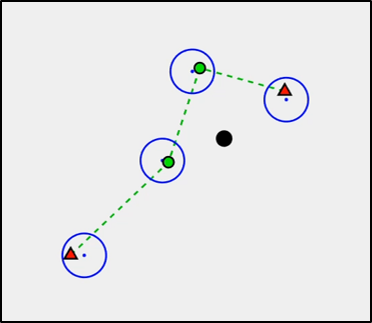}}\hfill \hspace{2cm} \\ \hspace{1cm}
  \subfloat[]{\includegraphics[width=0.23\linewidth]{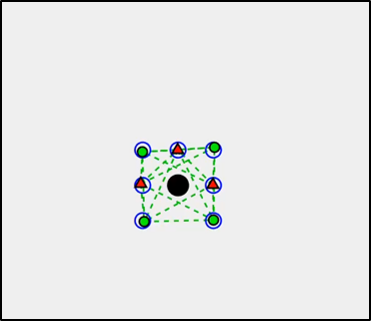}}\hfill
  \subfloat[]{\includegraphics[width=0.23\linewidth]{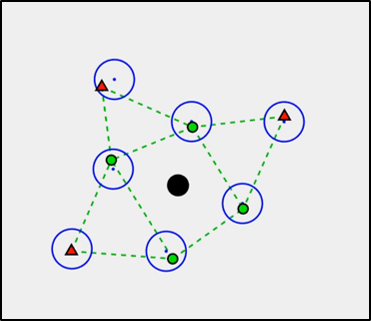}}\hfill
  \subfloat[]{\includegraphics[width=0.23\linewidth]{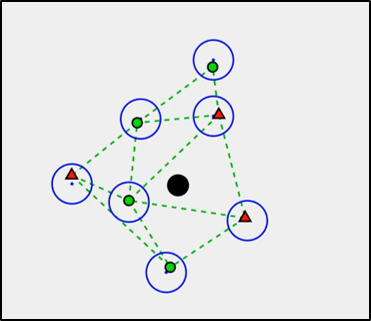}}\hfill \hspace{2cm} \\
  \caption{MATLAB simulations for various multi-robot system configurations. Snapshots of simulation runs where the follower robots implement our DCMU algorithm in three different configurations:(a)-(c), (d)-(f) and (g)-(i). The simulation videos can be viewed online at \url{https://youtu.be/SbE-ejQ_zm8}.}
  \label{fig:matlab-multi-robot-snapshots}
\end{figure}

\begin{figure}[t!]
  \centering
  \subfloat{\includegraphics[width=0.6\linewidth]{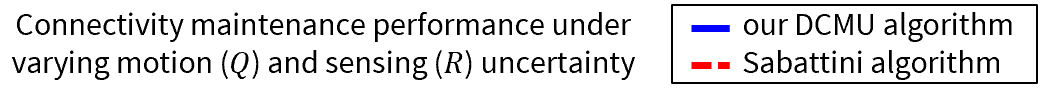}}\addtocounter{subfigure}{-1}\hfill \\ \vspace{0.1cm}
  \subfloat{\includegraphics[width=0.1\linewidth]{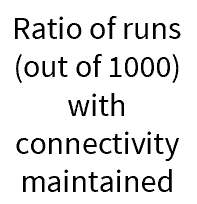}\vspace{1.6cm}}\addtocounter{subfigure}{-1}\hspace{-0.4cm}\hfill
  \subfloat[]{\includegraphics[width=0.27\linewidth]{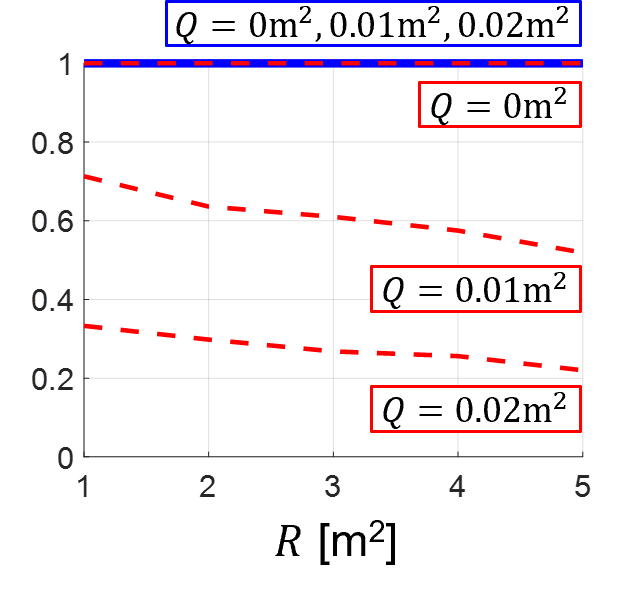}}\hspace{-0cm}\hfill
  \subfloat[]{\includegraphics[width=0.27\linewidth]{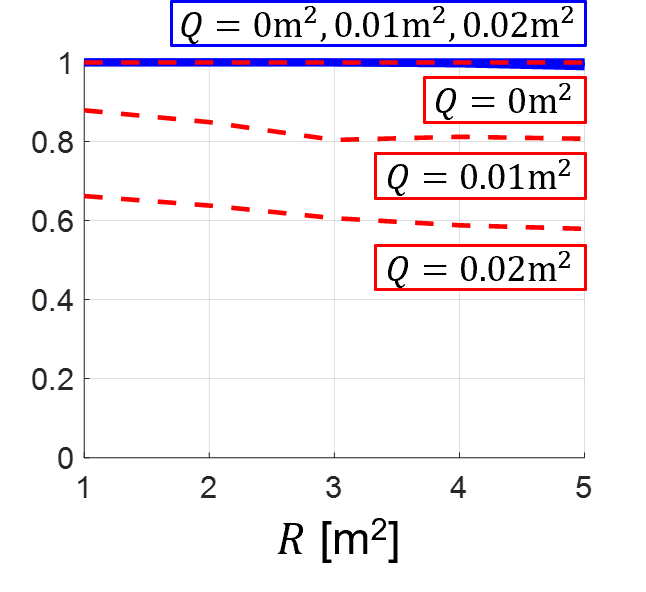}}\hfill
  \subfloat[]{\includegraphics[width=0.27\linewidth]{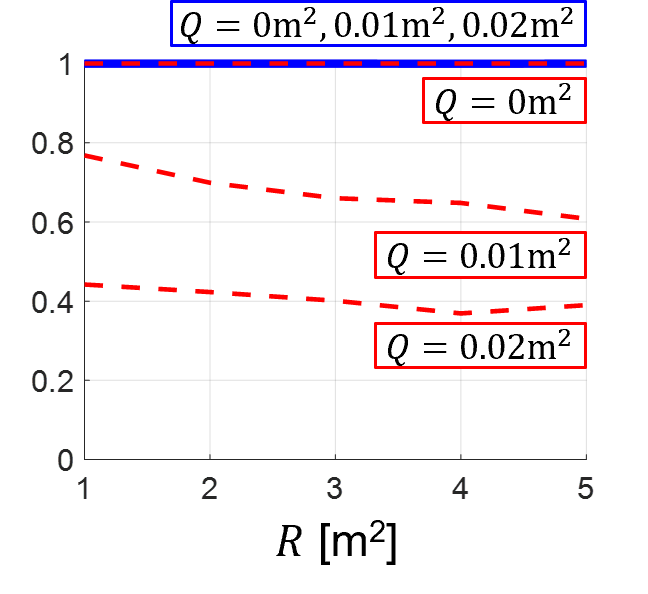}}\hspace{0.6cm}\hfill
  \caption{Quantitative comparisons for the multi-robot MATLAB simulations shown in Fig~\ref{fig:matlab-multi-robot-snapshots}. Our DCMU algorithm performs better than the Sabattini algorithm~\citep{Sabattini2013} under increasing motion and sensing uncertainties for the three multi-robot configurations.}
  \label{fig:matlab-multi-robot-quantitative}
\end{figure}

We then compare our algorithm with the Sabattini algorithm on different multi-robot system configurations as shown in Fig.~\ref{fig:matlab-multi-robot-snapshots}. Snapshots of a single simulation run with follower robots using our DCMU algorithm are shown for three different configurations: Fig.~\ref{fig:matlab-multi-robot-snapshots}(a)-(c), Fig.~\ref{fig:matlab-multi-robot-snapshots}(d)-(f) and Fig.~\ref{fig:matlab-multi-robot-snapshots}(g)-(i). The nominal trajectories for the leader robots cover distances of $\SI{80}{\meter}$ in $\SI{80}{\second}$, $\SI{60}{\meter}$ in $\SI{60}{\second}$, and $\SI{60}{\meter}$ in $\SI{60}{\second}$ respectively for the three configurations. Note that the follower robots are not assigned to follow specific leader robots. Instead, our DCMU algorithm computes nominal control inputs for the  follower robots directly based on each robot’s estimate of the system connectivity. Thus, in Fig.~\ref{fig:matlab-multi-robot-snapshots} we observe that the follower robots rearrange themselves in order to maintain connectivity within the system while accounting for the deviations arising due to motion and sensing uncertainties. Fig.~\ref{fig:matlab-multi-robot-quantitative} quantitatively compares the connectivity maintenance performance of our DCMU algorithm and the Sabattini algorithm for $1000$ simulation runs of each of the three multi-robot configurations. Similar to the two-robot case, we observe that our algorithm maintains $\bar{\eigenvalue}_2 > \AClimit$ throughout the run for all $1000$ runs, whereas the ratio of runs for which Sabattini algorithm maintains $\bar{\eigenvalue}_2 > \AClimit$ throughout the run sharply decreases as the amount of motion and sensing uncertainties increase. These simulations validate the applicability of our algorithm towards maintaining connectivity in the system under robot motion and sensing uncertainties.


\subsection{AirSim simulations}
\label{subsec:airsim-sims}

\begin{figure}[t!]
  \centering
  \hspace{-0.29cm} \subfloat{\includegraphics[width=0.25\linewidth]{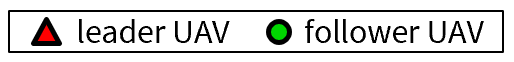}}\addtocounter{subfigure}{-1}\hfill \\ \hspace{1cm}
  \subfloat[]{\includegraphics[width=0.24\linewidth]{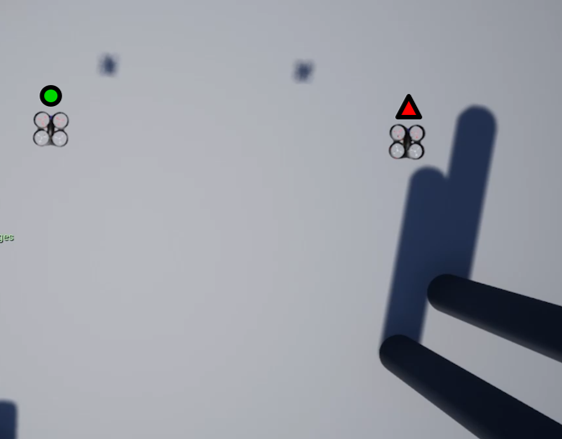}}\hfill
  \subfloat[]{\includegraphics[width=0.24\linewidth]{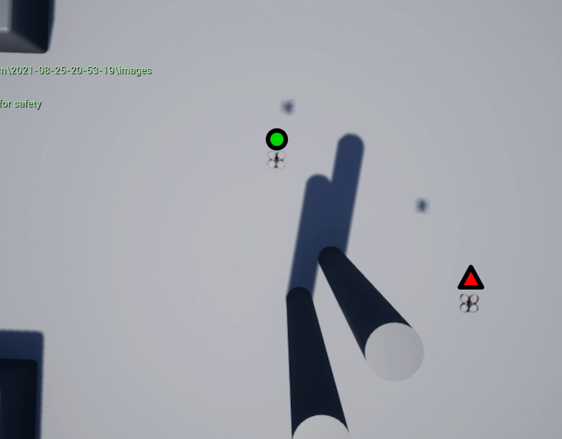}}\hfill
  \subfloat[]{\includegraphics[width=0.24\linewidth]{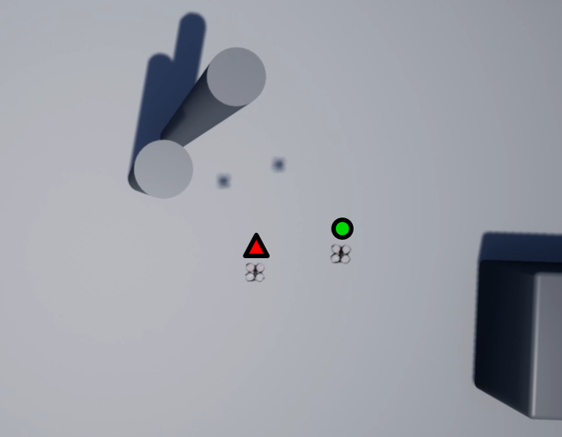}}\hfill \hspace{1cm} \\ \hspace{1cm}
  \subfloat[]{\includegraphics[width=0.24\linewidth]{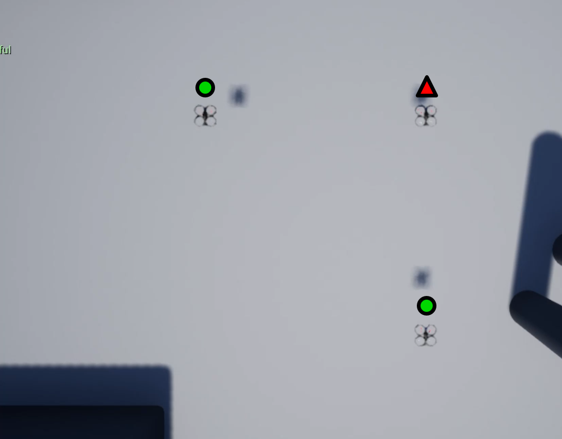}}\hfill
  \subfloat[]{\includegraphics[width=0.24\linewidth]{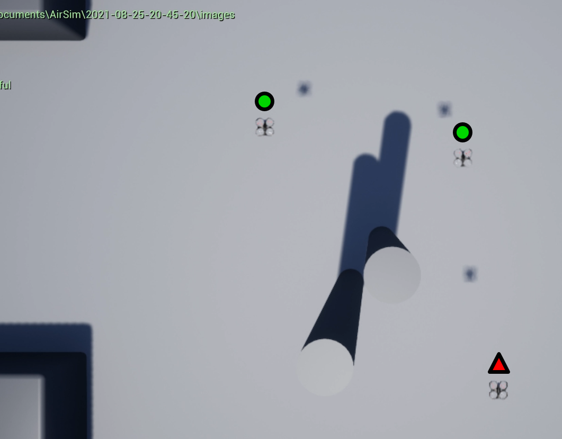}}\hfill
  \subfloat[]{\includegraphics[width=0.24\linewidth]{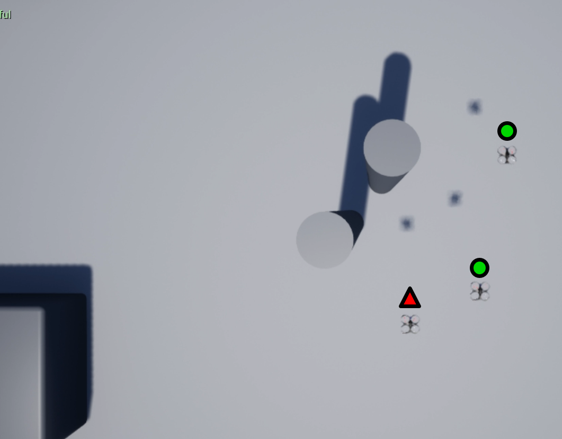}}\hfill \hspace{1cm} \\ \hspace{1cm}
  \subfloat[]{\includegraphics[width=0.24\linewidth]{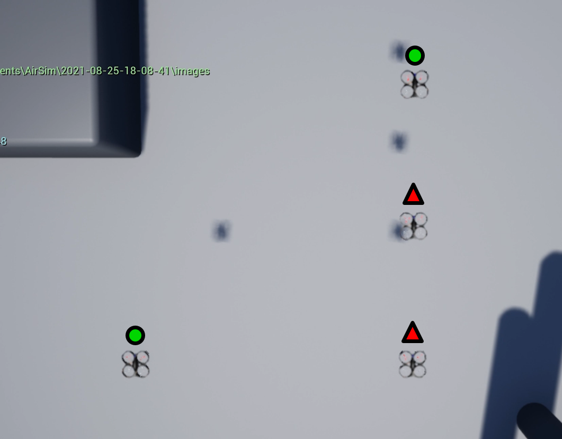}}\hfill
  \subfloat[]{\includegraphics[width=0.24\linewidth]{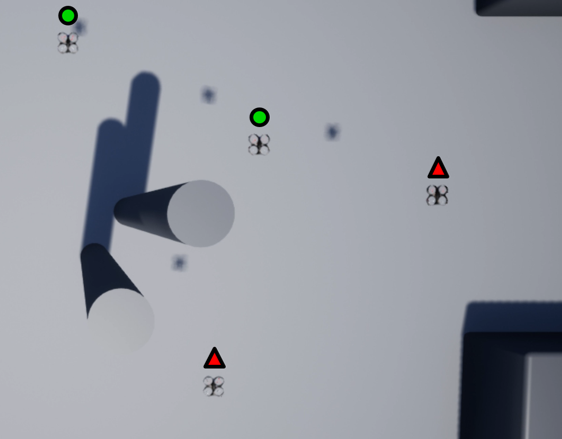}}\hfill
  \subfloat[]{\includegraphics[width=0.24\linewidth]{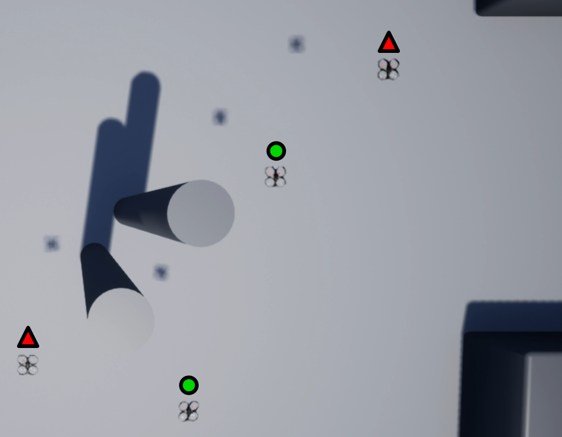}}\hfill \hspace{1cm} \\
  \caption{AirSim simulations for various multi-UAV system configurations. Snapshots of simulation runs where the follower UAVs implement our DCMU algorithm in three different configurations:(a)-(c), (d)-(f) and (g)-(i). The simulation videos can be viewed online at \url{https://youtu.be/SbE-ejQ_zm8}.}
  \label{fig:airsim-results-snapshots}
\end{figure}

\begin{figure}[t!]
  \centering
  \hspace{1cm}
  \subfloat[]{\includegraphics[width=0.28\linewidth]{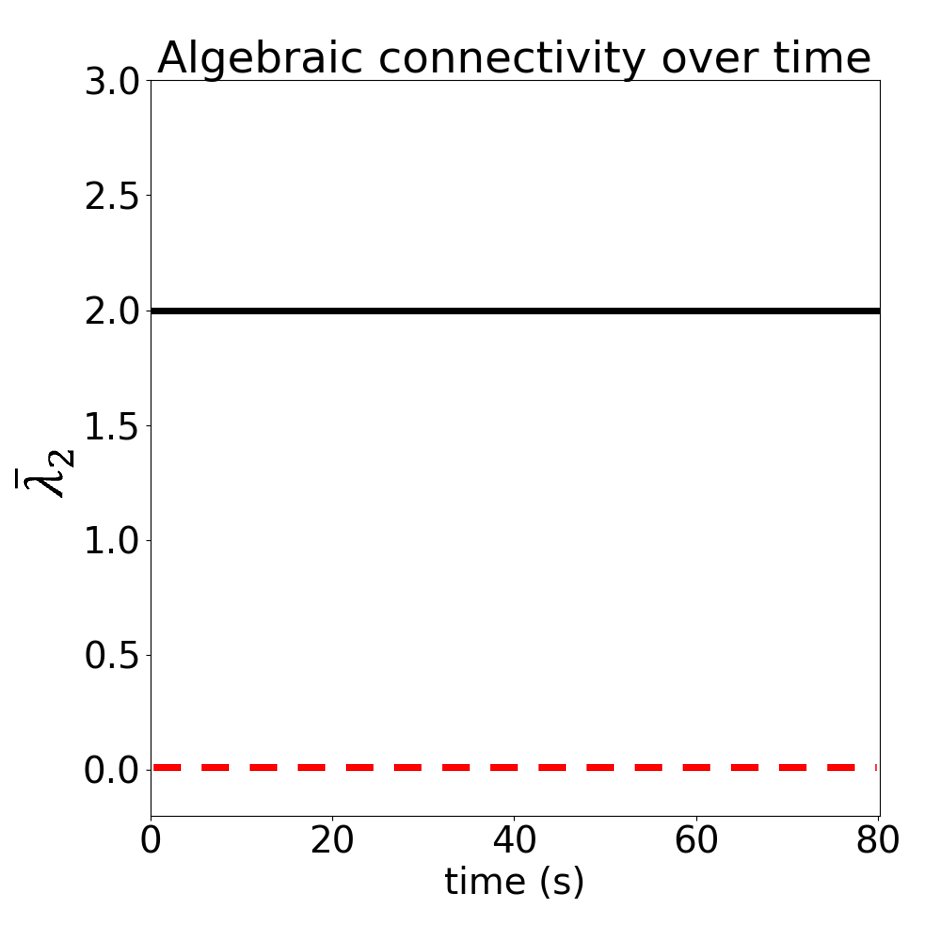}}\hfill
  \subfloat[]{\includegraphics[width=0.28\linewidth]{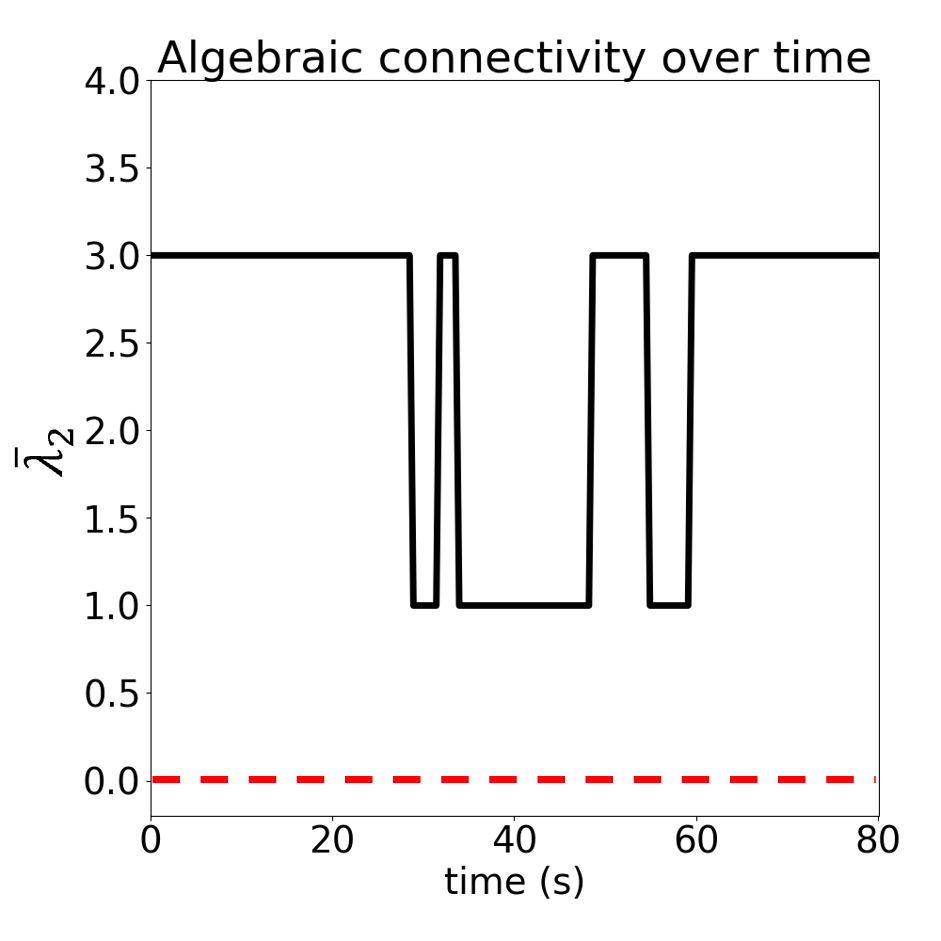}}\hfill
  \subfloat[]{\includegraphics[width=0.28\linewidth]{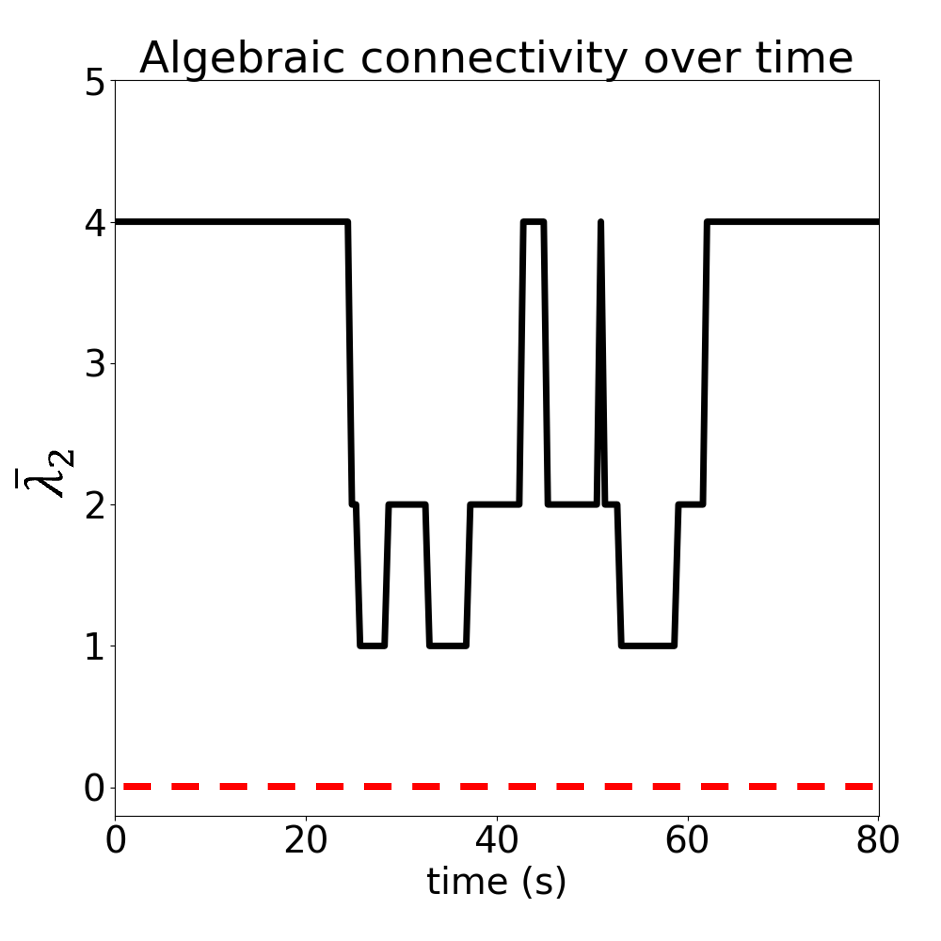}}\hfill \hspace{1cm}
  \caption{Algebraic connectivity of the multi-UAV AirSim simulations shown in Fig.~\ref{fig:airsim-results-snapshots}. Our DCMU algorithm maintains $\bar{\eigenvalue}_2$ (black lines) above the lower limit $\AClimit$ (red-dashed lines).\vspace{-0.4cm}}
  \label{fig:airsim-results-AC}
\end{figure}

For the AirSim simulations, we validate our algorithm on multi-robot systems comprising unmanned aerial vehicles (UAVs). The motion of the UAVs is modeled using Equation~(\ref{eqn:motion-model}), while sensing uncertainties are introduced according to Equation~(\ref{eqn:sensing-model}). We show snapshots with the follower UAVs using our DCMU algorithm for three different configurations: Fig.~\ref{fig:airsim-results-snapshots}(a)-(c), Fig.~\ref{fig:airsim-results-snapshots}(d)-(f) and Fig.~\ref{fig:airsim-results-snapshots}(g)-(i). The nominal trajectories for the leader robots cover distances of $\SI{105}{\meter}$ in $\SI{105}{\second}$, $\SI{180}{\meter}$ in $\SI{180}{\second}$, and $\SI{45}{\meter}$ in $\SI{90}{\second}$ respectively for the three configurations. The main objective of the snapshots in Fig.~\ref{fig:airsim-results-snapshots} is to demonstrate that our DCMU algorithm is able to compute nominal trajectories for the follower UAVs such that connectivity is maintained for different multi-UAV configurations while the leader UAVs follow their predefined nominal trajectories. Fig.~\ref{fig:airsim-results-AC} shows how the algebraic connectivity $\bar{\eigenvalue}_2$ is maintained above the lower limit $\AClimit$ throughout the simulation for the different configurations. Note that the AirSim simulations are executed in \textit{real-time}, thus validating the real-time applicability of our DCMU algorithm.

\section{Conclusion}\label{sec:conclusion}

In this paper we presented a Decentralized Connectivity Maintenance algorithm that accounted for robot motion and sensing Uncertainties (DCMU). We first defined a weighted graph to account for these uncertainties along with constraints such as a maximum communication range, line-of-sight communication and collision avoidance. Next, we designed a decentralized gradient-based controller by deriving the gradients of our weighted graph edge weights. Finally, we validated the connectivity maintenance performance of our algorithm on two simulation setups: MATLAB and AirSim. We quantitatively compared our DCMU algorithm with previous related work~\citep{Sabattini2013} and demonstrated that our algorithm performed better under motion and sensing uncertainties. Additionally, we also demonstrated the connectivity maintenance performance of our algorithm on AirSim, which is a higher-fidelity simulator compared to MATLAB.

Possible directions for future work include exploring alternate geometric representations that reduce the conservatism in our weighted graph\new{, evaluating connectivity maintenance performance for a range of graph parameter values,} and accounting for more complex robot motion models such as ground robots with nonholonomic constraints. Additionally, we plan to evaluate our DCMU algorithm on real hardware and analyze the effects of communication latency on the system connectivity maintenance.

\printbibliography[heading=bibintoc, title={References}]

\end{document}